\documentclass[manuscript,screen]{acmart}
\usepackage{tikz}
\usepackage{adjustbox}
\usetikzlibrary{positioning,calc,backgrounds,patterns,arrows.meta,bending,shapes.geometric}

\usepackage{booktabs}      
\usepackage{multirow}      
\usepackage{array}         
\usepackage{tabularx}      
\usepackage{makecell}      
\usepackage{rotating}      
\usepackage{pdflscape}     
\usepackage{enumitem}      
\usepackage{amsmath}       

\usepackage{amssymb}       
\usepackage{algorithm}
\usepackage{algpseudocode}
\usepackage[acronym]{glossaries}

\usepackage{xcolor}

\definecolor{cStatic}{RGB}{68,119,170}
\definecolor{cDiscrete}{RGB}{204,187,68}
\definecolor{cContinuous}{RGB}{68,153,102}
\definecolor{cImplicit}{RGB}{153,68,153}
\definecolor{cHybrid}{RGB}{204,68,68}

\definecolor{cGrTeal}{RGB}{68,170,187}
\definecolor{cGrOlive}{RGB}{170,187,102}
\definecolor{cGrRose}{RGB}{204,119,170}
\definecolor{cGrCoral}{RGB}{238,119,102}
\definecolor{cGrGold}{RGB}{204,170,102}

\makeglossaries
\newacronym{ann}{ANN}{Artificial Neural Network}
\newacronym{ai}{AI}{Artificial Intelligence}
\newacronym{rnn}{RNN}{Recurrent Neural Network}
\newacronym{lstm}{LSTM}{Long Short-Term Memory}
\newacronym{gru}{GRU}{Gated Recurrent Unit}
\newacronym{snn}{SNN}{Spiking Neural Network}
\newacronym{node}{NODE}{Neural Ordinary Differential Equation}
\newacronym{ssm}{SSM}{State-Space Model}
\newacronym{stdp}{STDP}{Spike-Timing-Dependent Plasticity}
\newacronym{deq}{DEQ}{Deep Equilibrium Model}
\newacronym{lif}{LIF}{Leaky Integrate-and-Fire}
\newacronym{mlp}{MLP}{Multilayer Perceptron}
\newacronym{cnn}{CNN}{Convolutional Neural Network}
\newacronym{resnet}{ResNet}{Residual Network}
\newacronym{pinn}{PINN}{Physics-Informed Neural Network}
\newacronym{esn}{ESN}{Echo State Network}
\newacronym{lsm}{LSM}{Liquid State Machine}
\newacronym{htm}{HTM}{Hierarchical Temporal Memory}
\newacronym{rbm}{RBM}{Restricted Boltzmann Machine}
\newacronym{dbn}{DBN}{Deep Belief Network}
\newacronym{fep}{FEP}{Free-Energy Principle}
\newacronym{aif}{AIF}{Active Inference}
\newacronym{bp}{BP}{Backpropagation}
\newacronym{bptt}{BPTT}{Backpropagation Through Time}
\newacronym{ctrnn}{CTRNN}{Continuous-Time Recurrent Neural Network}
\newacronym{pde}{PDE}{Partial Differential Equation}
\newacronym{gpu}{GPU}{Graphics Processing Unit}
\newacronym{tpu}{TPU}{Tensor Processing Unit}
\newacronym{cpu}{CPU}{Central Processing Unit}

\setcopyright{none}
\copyrightyear{2026}
\acmYear{2026}
\acmJournal{CSUR}

\AtBeginDocument{%
  }

\title{The Forward--Backward Disconnect: State Dynamics, Credit Assignment, and Biological Grounding in Neural Computation}

\author{Hadi Al Mubasher}
\email{hma154@aub.edu.lb}
\orcid{0000-0001-7511-2910}
\affiliation{%
  \institution{Department of Electrical and Computer Engineering,
  Maroun Semaan Faculty of Engineering and Architecture,
  American University of Beirut}
  \city{Beirut}
  \country{Lebanon}
}

\author{Mariette Awad}
\email{mariette.awad@aub.edu.lb}
\affiliation{%
  \institution{Department of Electrical and Computer Engineering,
  Maroun Semaan Faculty of Engineering and Architecture,
  American University of Beirut}
  \city{Beirut}
  \country{Lebanon}
}

\renewcommand{\shortauthors}{Al Mubasher et al.}



\begin{document}

\begin{abstract}
A recurring pattern in neural computation is the reintroduction of dynamical and biological structure into models that were originally simplified for scalable optimization. Early feedforward networks, from McCulloch--Pitts units through the perceptron and its multilayer descendants, reduced biological neurons to threshold or rate-like summation units. That abstraction proved structurally compatible with global-gradient training at scale. Since then, the forward side of neural computation has diversified: modern architectures carry state through recurrent updates, retrieve from long contexts through attention and associative memory, compress histories through structured state-space dynamics, evolve along continuous-time trajectories, settle to implicit equilibria, and communicate through sparse event-driven spikes.
 
Training has diversified less. The mechanisms that make these systems scalable remain concentrated around backpropagation, backpropagation through time, adjoint methods, implicit differentiation, and surrogate-gradient variants. This survey names the resulting asymmetry between diversified forward dynamics and concentrated credit assignment the \emph{forward--backward disconnect}, and develops a taxonomy for analyzing it across neural model families.
 
The taxonomy organizes models along three coupled axes: state-dynamics structure, credit-assignment mechanism, and biological grounding. Biological grounding is split into separate forward and learning dimensions because a model may preserve biologically motivated dynamics while relying on a non-biological training rule. The unit of analysis is therefore the architecture--learning configuration rather than the architecture name alone: the same forward operator can occupy different taxonomic positions depending on whether it is trained by global gradients, approximate gradients, local plasticity, or energy-based objectives.
 
Across static, recurrent, attention-based, state-space, continuous-time, implicit, spiking, biologically plausible, and neuromorphic families, the taxonomy reveals a consistent pattern: forward computation has diversified across state-dynamics classes, while the highest demonstrated scales remain concentrated in global or closely gradient-derived and error-propagation mechanisms. Biology serves as a diagnostic constraint rather than as metaphor, exposing which mechanisms are preserved, discarded, or replaced by engineering substitutes. Our synthesis suggests that closing the disconnect requires configuration-level alignment among state dynamics, credit assignment, and computational substrate.
\end{abstract}
 
\maketitle
 
\section{Introduction}
\label{sec:introduction}
 
A central simplification behind modern deep learning is that neural computation can be made scalable by treating a network as a differentiable input--output map, an abstraction that allowed neural networks to scale across vision, language, speech, and scientific modeling \cite{lecun2015deep,goodfellow2016deep}. The forward side of neural computation has since moved beyond this simplification: modern architectures compute through recurrent hidden states, attention-based retrieval, structured state-space evolution, continuous-time trajectories, implicit equilibria, and event-driven spikes. The learning side has changed less; across many of these families, the mechanism that makes the model trainable remains backpropagation, backpropagation through time, adjoint differentiation, implicit differentiation, or a surrogate-gradient variant. This survey names the resulting asymmetry the \emph{forward--backward disconnect} and treats it as a structural mismatch among three objects usually studied separately: the dynamics used to compute, the learning rule used to assign credit, and the hardware substrate on which both are implemented.
 
The static abstraction remains the baseline from which the taxonomy departs:
\begin{equation}
    \mathbf{y} = f_{\theta}(\mathbf{x}),
    \label{eq:intro_static_mapping}
\end{equation}
with \(\mathbf{x}\) the input, \(\mathbf{y}\) the output, and \(\theta\) the learnable parameters. This survey uses \emph{static} in the dynamical-systems sense: a memoryless input--output map with no persistent internal state carried across evaluations. \emph{Dynamical} correspondingly refers to a computation whose internal state evolves under a state-transition rule and influences future outputs. Feedforward networks (\gls{mlp}, \gls{cnn}, \gls{resnet}) are static in this sense; recurrent, continuous-time, spiking, and implicit-equilibrium models are dynamical. The usage differs from ML conventions in which ``static'' may refer to a fixed computation graph as opposed to define-by-run execution, or to parameters held frozen during inference or fine-tuning.
 
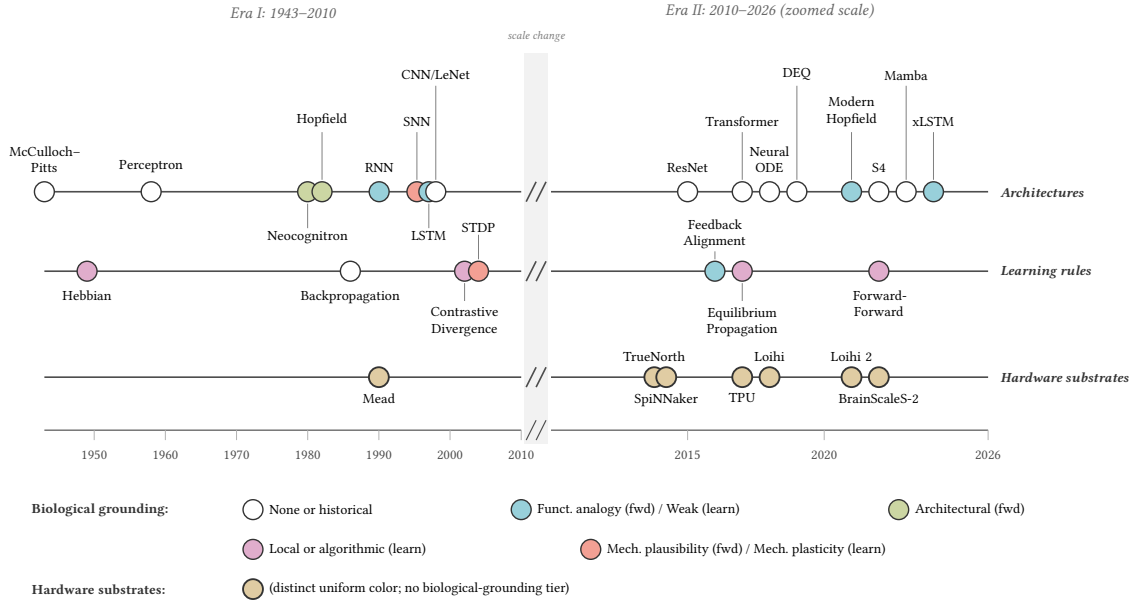
\begin{figure*}[t]
\centering
\begin{adjustbox}{max width=\textwidth}
\begin{tikzpicture}[
  x=0.15cm, y=1.0cm,
  axis style/.style = {line width=0.7pt, black!70},
  tickline/.style   = {line width=0.3pt, black!30},
  yearlabel/.style  = {font=\scriptsize, anchor=north, text=black!70},
  track label/.style = {font=\scriptsize\itshape\bfseries, anchor=west, text=black!80},
  segment label/.style = {font=\footnotesize\itshape, anchor=south, text=black!60},
  tl histinsp/.style       = {circle, draw=black!80, fill=white,         line width=0.6pt, minimum size=0.30cm, inner sep=0pt},
  tl funcanalogy/.style    = {circle, draw=black!80, fill=cGrTeal!55,    line width=0.6pt, minimum size=0.30cm, inner sep=0pt},
  tl architectural/.style  = {circle, draw=black!80, fill=cGrOlive!60,   line width=0.6pt, minimum size=0.30cm, inner sep=0pt},
  tl algorithmic/.style    = {circle, draw=black!80, fill=cGrRose!60,    line width=0.6pt, minimum size=0.30cm, inner sep=0pt},
  tl mechplausible/.style  = {circle, draw=black!80, fill=cGrCoral!70,   line width=0.6pt, minimum size=0.30cm, inner sep=0pt},
  tl hardware/.style       = {circle, draw=black!80, fill=cGrGold!60,    line width=0.8pt, minimum size=0.30cm, inner sep=0pt},
  ulab/.style = {font=\scriptsize, align=center, anchor=south, inner sep=1pt},
  llab/.style = {font=\scriptsize, align=center, anchor=north, inner sep=1pt},
  leader/.style     = {line width=0.3pt, black!50}
]
 
\useasboundingbox (-7, -5.9) rectangle (108, 4.5);
 
\fill[black!5] (48.3, 2.7) rectangle (50.7, -3.2);
 
\node[font=\tiny\itshape, text=black!50, anchor=center] at (49.5, 2.95) {scale change};
 
\node[segment label] at (24, 3.10) {Era I: 1943--2010};
\node[segment label] at (73, 3.10) {Era II: 2010--2026 (zoomed scale)};
 
\draw[axis style] (0, 0.6) -- (48, 0.6);
\draw[axis style] (48.5, 0.45) -- (49.5, 0.75);
\draw[axis style] (49.5, 0.45) -- (50.5, 0.75);
\draw[axis style] (51, 0.6) -- (95, 0.6);
\node[track label] at (95.5, 0.6) {Architectures};
 
\draw[axis style] (0, -0.6) -- (48, -0.6);
\draw[axis style] (48.5, -0.75) -- (49.5, -0.45);
\draw[axis style] (49.5, -0.75) -- (50.5, -0.45);
\draw[axis style] (51, -0.6) -- (95, -0.6);
\node[track label] at (95.5, -0.6) {Learning rules};
 
\draw[axis style, line width=0.5pt] (0, -3.0) -- (48, -3.0);
\draw[axis style, line width=0.5pt] (51, -3.0) -- (95, -3.0);
\foreach \y in {1950,1960,1970,1980,1990,2000,2010} {
  \pgfmathsetmacro{\xpos}{(\y - 1943) * 48 / 67}
  \draw[tickline] (\xpos, -3.0) -- (\xpos, -3.15);
  \node[yearlabel, anchor=north] at (\xpos, -3.20) {\y};
}
\foreach \y in {2015,2020,2026} {
  \pgfmathsetmacro{\xpos}{51 + (\y - 2010) * 44 / 16}
  \draw[tickline] (\xpos, -3.0) -- (\xpos, -3.15);
  \node[yearlabel, anchor=north] at (\xpos, -3.20) {\y};
}
\draw[axis style, line width=0.5pt] (48.5, -3.15) -- (49.5, -2.85);
\draw[axis style, line width=0.5pt] (49.5, -3.15) -- (50.5, -2.85);
 
\node[tl histinsp] at (0, 0.6) {};
\node[ulab, text width=1.6cm] at (0, 0.85) {McCulloch--\\Pitts};
\node[tl histinsp] at (10.75, 0.6) {};
\node[ulab, text width=1.4cm] at (10.75, 0.85) {Perceptron};
\node[tl architectural] at (26.5, 0.6) {};
\draw[leader] (26.5, 0.42) -- (26.5, 0.05);
\node[llab, text width=1.6cm] at (26.5, 0.05) {Neocognitron};
\node[tl architectural] at (27.95, 0.6) {};
\draw[leader] (27.95, 0.78) -- (27.95, 1.50);
\node[ulab, text width=1.2cm] at (27.95, 1.55) {Hopfield};
\node[tl funcanalogy] at (33.7, 0.6) {};
\node[ulab, text width=1.4cm] at (33.7, 0.85) {RNN};
\node[tl mechplausible] at (37.5, 0.6) {};
\draw[leader] (37.5, 0.78) -- (37.5, 1.50);
\node[ulab, text width=1.2cm] at (37.5, 1.55) {SNN};
\node[tl funcanalogy] at (38.7, 0.6) {};
\draw[leader] (38.7, 0.42) -- (38.7, 0.05);
\node[llab, text width=1.0cm] at (38.7, 0.05) {LSTM};
\node[tl histinsp] at (39.4, 0.6) {};
\draw[leader] (39.4, 0.78) -- (39.4, 2.20);
\node[ulab, text width=1.6cm] at (39.4, 2.25) {CNN/LeNet};
 
\node[tl histinsp] at (64.75, 0.6) {};
\node[ulab, text width=1.4cm] at (64.75, 0.85) {ResNet};
\node[tl histinsp] at (73, 0.6) {};
\node[ulab, text width=1.4cm] at (73, 0.85) {Neural\\ODE};
\node[tl histinsp] at (84, 0.6) {};
\node[ulab, text width=1.0cm] at (84, 0.85) {S4};
\node[tl histinsp] at (70.25, 0.6) {};
\draw[leader] (70.25, 0.78) -- (70.25, 1.5);
\node[ulab, text width=1.5cm] at (70.25, 1.55) {Transformer};
\node[tl funcanalogy] at (81.25, 0.6) {};
\draw[leader] (81.25, 0.78) -- (81.25, 1.5);
\node[ulab, text width=1.5cm] at (81.25, 1.55) {Modern\\Hopfield};
\node[tl funcanalogy] at (89.5, 0.6) {};
\draw[leader] (89.5, 0.78) -- (89.5, 1.5);
\node[ulab, text width=1.2cm] at (89.5, 1.55) {xLSTM};
\node[tl histinsp] at (75.75, 0.6) {};
\draw[leader] (75.75, 0.78) -- (75.75, 2.20);
\node[ulab, text width=1.0cm] at (75.75, 2.25) {DEQ};
\node[tl histinsp] at (86.75, 0.6) {};
\draw[leader] (86.75, 0.78) -- (86.75, 2.20);
\node[ulab, text width=1.2cm] at (86.75, 2.25) {Mamba};
 
\node[tl algorithmic] at (4.3, -0.6) {};
\node[llab, text width=1.4cm] at (4.3, -0.85) {Hebbian};
\node[tl histinsp] at (30.8, -0.6) {};
\node[llab, text width=2.0cm] at (30.8, -0.85) {Backpropagation};
\node[tl algorithmic] at (42.3, -0.6) {};
\draw[leader] (42.3, -0.78) -- (42.3, -1.05);
\node[llab, text width=1.6cm] at (42.3, -1.10) {Contrastive\\Divergence};
\node[tl mechplausible] at (43.7, -0.6) {};
\draw[leader] (43.7, -0.42) -- (43.7, -0.05);
\node[ulab, text width=1.0cm] at (43.7, -0.05) {STDP};
 
\node[tl funcanalogy] at (67.5, -0.6) {};
\draw[leader] (67.5, -0.42) -- (67.5, -0.30);
\node[ulab, text width=1.6cm] at (67.5, -0.30) {Feedback\\Alignment};
\node[tl algorithmic] at (70.25, -0.6) {};
\draw[leader] (70.25, -0.78) -- (70.25, -1.05);
\node[llab, text width=1.6cm] at (70.25, -1.10) {Equilibrium\\Propagation};
\node[tl algorithmic] at (84, -0.6) {};
\node[llab, text width=1.5cm] at (84, -0.85) {Forward-\\Forward};
 
\draw[axis style] (0, -2.2) -- (48, -2.2);
\draw[axis style] (48.5, -2.35) -- (49.5, -2.05);
\draw[axis style] (49.5, -2.35) -- (50.5, -2.05);
\draw[axis style] (51, -2.2) -- (95, -2.2);
\node[track label] at (95.5, -2.2) {Hardware substrates};
 
\node[tl hardware] at (33.67, -2.2) {};
\node[llab, text width=1.2cm] at (33.67, -2.40) {Mead};
 
\node[tl hardware] at (61.4, -2.2) {};
\node[ulab, text width=1.4cm] at (61.4, -2.00) {TrueNorth};
\node[tl hardware] at (62.6, -2.2) {};
\node[llab, text width=1.4cm] at (62.6, -2.40) {SpiNNaker};
\node[tl hardware] at (70.25, -2.2) {};
\node[llab, text width=1.0cm] at (70.25, -2.40) {TPU};
\node[tl hardware] at (73, -2.2) {};
\node[ulab, text width=1.0cm] at (73, -2.00) {Loihi};
\node[tl hardware] at (81.25, -2.2) {};
\node[ulab, text width=1.2cm] at (81.25, -2.00) {Loihi 2};
\node[tl hardware] at (84, -2.2) {};
\node[llab, text width=2.0cm] at (84, -2.40) {BrainScaleS-2};
 
\begin{scope}[shift={(0, -4.2)}]
  \node[font=\scriptsize\bfseries, anchor=west, text=black!80] at (-2, 0) {Biological grounding:};
  \node[tl histinsp]      at (21, 0) {};
  \node[anchor=west, font=\scriptsize] at (21.9, 0) {None or historical};
  \node[tl funcanalogy]   at (48, 0) {};
  \node[anchor=west, font=\scriptsize] at (48.9, 0) {Funct.\ analogy (fwd) / Weak (learn)};
  \node[tl architectural] at (86, 0) {};
  \node[anchor=west, font=\scriptsize] at (86.9, 0) {Architectural (fwd)};
\end{scope}
 
\begin{scope}[shift={(0, -4.8)}]
  \node[tl algorithmic]   at (21, 0) {};
  \node[anchor=west, font=\scriptsize] at (21.9, 0) {Local or algorithmic (learn)};
  \node[tl mechplausible] at (55, 0) {};
  \node[anchor=west, font=\scriptsize] at (55.9, 0) {Mech.\ plausibility (fwd) / Mech.\ plasticity (learn)};
\end{scope}
 
\begin{scope}[shift={(0, -5.4)}]
  \node[font=\scriptsize\bfseries, anchor=west, text=black!80] at (-2, 0) {Hardware substrates:};
  \node[tl hardware]      at (21, 0) {};
  \node[anchor=west, font=\scriptsize] at (21.9, 0) {(distinct uniform color; no biological-grounding tier)};
\end{scope}
 
\end{tikzpicture}
\end{adjustbox}
\caption{Timeline of selected architectural, learning-rule, and hardware-substrate milestones in neural computation, 1943--2026. Marker colors encode biological grounding under the split-axis scheme of Section~\ref{sec:taxonomy_bio}; hardware uses a distinct gold marker because substrates do not map onto grounding tiers. The post-2010 segment is horizontally zoomed to expose the greater density of recent milestones. The figure provides illustrative historical orientation rather than quantitative evidence; Figure~\ref{fig:master_taxonomy} and Section~\ref{sec:open_synthesis} establish the forward--backward asymmetry categorically.}
\Description{A three-track timeline from 1943 to 2026. The upper track shows selected architectural milestones in neural computation; the middle track shows selected learning-rule and credit-assignment milestones; the lower track shows selected hardware-substrate milestones, including neuromorphic computing introduced by Mead in 1990, the digital neuromorphic chip TrueNorth and the parallel ARM-core machine SpiNNaker in 2014, the systolic-array Tensor Processing Unit in 2017, the Loihi chip in 2018 with on-chip plasticity, Loihi 2 in 2021, and the mixed-signal BrainScaleS-2 in 2022. A visual axis break at 2010 separates a pre-2010 segment from a post-2010 segment drawn at a horizontally zoomed scale. Architecture-track and learning-track marker color encodes biological grounding under a shared color scheme: white for none or historical inspiration; teal for functional analogy on the forward axis and weak on the learning axis; olive for architectural constraint; rose for local or algorithmic; and coral for mechanistic plausibility on the forward axis and mechanistic plasticity on the learning axis. Hardware-track markers are filled gold because compute substrates do not map onto biological-grounding tiers.}
\label{fig:timeline}
\end{figure*}

Neural-network history is a gradual reintroduction of dynamical structure that the static abstraction removed. The earliest models, from McCulloch and Pitts \cite{mcculloch1943logical} through Rosenblatt's perceptron \cite{rosenblatt1958perceptron}, bridged biological inspiration and mathematical computation but discarded the membrane-potential dynamics, recurrent interaction, and spike-based signaling of real neural systems. Successive developments restored these features unevenly: \glspl{rnn} with discrete-time recurrence \cite{elman1990finding}; \glspl{lstm} and \glspl{gru} with gated memory \cite{hochreiter1997long,cho2014learning}; \glspl{snn} with event-driven communication \cite{maass1997networks}; \glspl{node} with continuous trajectories \cite{chen2018neural}; structured \glspl{ssm} with explicit state evolution for long-range sequence modeling \cite{gu2022efficiently,gu2023mamba}. In parallel, biologically motivated learning rules (Hebbian plasticity, \gls{stdp}, predictive coding, feedback alignment, equilibrium propagation) questioned whether backpropagation is the only plausible credit-assignment mechanism for deep networks \cite{hebb1949organization,bi1998synaptic,rao1999predictive,lillicrap2016random,scellier2017equilibrium}. Figure~\ref{fig:timeline} arranges these milestones on a three-track timeline: the architectural track shows a dense post-2015 cluster (ResNet, Transformer, Neural ODE, DEQ, Modern Hopfield, S4, Mamba, xLSTM); the hardware track shows a parallel substrate cluster (\gls{tpu}, Loihi, Loihi 2, BrainScaleS-2); the learning-rule track shows only sparse alternatives (feedback alignment, equilibrium propagation, Forward-Forward \cite{hinton2022forward}). This chronological asymmetry complements the categorical view of Figure~\ref{fig:master_taxonomy}.
 
The objective of this survey is taxonomic reconstruction rather than chronological history. Model families are classified along two coupled axes: the form of state dynamics (no persistent state, discrete-time recurrence, continuous-time trajectories, implicit fixed-point states, or hybrid event-driven dynamics) and the learning or credit-assignment mechanism (global supervised gradients, approximate or implicit gradients, local plasticity, energy-based objectives, or unsupervised and generative learning). Biological grounding overlays both axes because it depends jointly on the forward architecture and the learning rule. A model family is included when it instantiates nontrivial state dynamics or is substantively grounded in neuroscience in a way that constrains design rather than merely motivates by analogy; the full scope rationale is given in Sections~\ref{sec:foundations_joint} and~\ref{sec:taxonomy_scope}.
 
Two networks with identical forward dynamics can occupy completely different positions in this taxonomy depending only on how they are trained: the joint axes of forward dynamics and credit assignment turn out to be more loosely coupled than the joint axes of biological dynamics and synaptic plasticity in the systems that motivated neural networks in the first place. The disconnect is not held in place by mathematical preference alone; it is materially reinforced by the compute substrate. Mainstream accelerators remain disproportionately optimized for dense batched tensor operations, whereas sparse asynchronous local updates typically achieve poorer utilization and less favorable memory behavior, exemplifying the hardware lottery of \cite{hooker2021hardware} in which the prevailing learning rule is the one whose computational pattern best matches the available substrate. Section~\ref{sec:neuromorphic} returns to this as the substrate side of the disconnect.
 
The disconnect appears repeatedly across the survey. Spiking networks restore membrane-potential dynamics and event-driven communication, but their scalable variants rely on surrogate gradients rather than mechanistic plasticity; continuous-time and implicit models change the forward state representation but recover gradients through adjoint or implicit-differentiation machinery; neuromorphic hardware aligns naturally with sparse event-driven computation but does not by itself supply scalable credit assignment. Throughout, biology is treated as a discussant rather than metaphor: at each family we ask what computational problem biology solves, what the artificial model preserves, and what it ignores. This commitment shapes the split-axis biological-grounding scheme of Section~\ref{sec:taxonomy_bio}.
 
This survey makes four contributions. First, it organizes neural model families by state-dynamics structure rather than by architecture name alone. Second, it introduces a credit-assignment axis that separates global gradients, approximate or implicit gradients, local plasticity, and energy-based objectives. Third, it splits biological grounding into forward and learning dimensions, allowing cases such as surrogate-gradient and \gls{stdp}-trained \glspl{snn} to be distinguished. Fourth, it connects the algorithmic taxonomy to the substrate question, arguing that closing the forward--backward disconnect likely requires architecture--learning--hardware co-design.
 
The survey is organized in three parts. Sections~\ref{sec:foundations} and~\ref{sec:taxonomy} establish the biological and dynamical vocabulary and the joint state-dynamics/credit-assignment taxonomy with split biological-grounding axes. Sections~\ref{sec:static}--\ref{sec:spiking} traverse the taxonomy by column: static feedforward, discrete-time recurrent, attention and structured state-space, continuous-time recurrent, and hybrid event-driven systems. Sections~\ref{sec:learning} and~\ref{sec:neuromorphic} turn to biologically plausible learning and the neuromorphic substrate. Sections~\ref{sec:open_problems} and~\ref{sec:conclusion} synthesize open problems and conclude.

\section{Biological and Dynamical Foundations}
\label{sec:foundations}
 
Neural computation, both biological and artificial, can be characterized by the form of its state dynamics, and the taxonomy of Section~\ref{sec:taxonomy} rests on this view. We use \emph{state} in the dynamical-systems sense throughout: the set of variables that, together with the present input, determines future evolution without requiring the full past history. In neural models, this may include membrane voltage, channel-gating variables, synaptic conductances, adaptation traces, hidden vectors, or equilibrium variables. Notation conventions: \(\mathbf{x}\) denotes inputs (consistent with Eq.~\eqref{eq:intro_static_mapping}), \(\mathbf{h}\) hidden state variables in explicit-time models, and \(\mathbf{z}\) latent or implicit-state representations. At the single-neuron level, biophysical state variables retain their conventional names (\(V\) for membrane potential, gating variables, conductances).
 
\subsection{Biological Neurons and the Limits of the Point-Neuron Abstraction}
\label{sec:foundations_neurons}
 
A biological neuron consists of dendrites, soma, and axon: presynaptic spikes trigger neurotransmitter release, postsynaptic currents integrate across the dendritic tree, and the soma with axon initial segment determines whether integrated activity reaches threshold for an action potential \cite{dayan2001theoretical,gerstner2014neuronal}. At the biophysical level the neuron is a dynamical system whose membrane acts as a capacitor driven by ionic and synaptic currents; the Hodgkin--Huxley model \cite{hodgkin1952quantitative} captures this as coupled differential equations in which membrane potential and channel-gating variables co-evolve.
 
Most artificial networks reduce this to a point-neuron abstraction, collapsing synaptic integration to a weighted sum and producing an output through a nonlinear activation. The reduction makes large-scale learning tractable but discards the spatial, temporal, and biophysical state variables that make biological neurons dynamical systems. The most common reduced model that preserves temporal state is the \gls{lif} neuron:
\begin{equation}
\tau_m \frac{dV(t)}{dt} = -\bigl(V(t) - E_L\bigr) + R_m I(t),
\qquad
V(t^-) \ge V_{\mathrm{th}} \;\Rightarrow\; \text{spike}, \;\; V(t^+) \leftarrow V_{\mathrm{reset}},
\label{eq:lif_single}
\end{equation}
which combines continuous subthreshold integration with a threshold-and-reset event \cite{lapicque1907recherches,gerstner2002spiking,gerstner2014neuronal}. \gls{lif} models are hybrid and event-driven: subthreshold state evolves continuously while communication and reset occur at discrete spike events \cite{maass1997networks}. The point-neuron abstraction is itself incomplete: cortical pyramidal neurons process inputs across extended dendritic trees with local nonlinearities and compartment-specific mechanisms that substantially increase single-neuron computational capacity \cite{poirazi2003pyramidal,london2005dendritic,larkum2013cellular,stuart2015dendritic,beniaguev2021single}, and together with membrane-potential dynamics and threshold-triggered spike events these form the structural basis of the \emph{Hybrid event-driven} state-dynamics class of Section~\ref{sec:taxonomy_dynamics}.
 
\subsection{Rate, Temporal, and Population Coding}
\label{sec:foundations_coding}
 
Three coding schemes are relevant to the taxonomy; they are not mutually exclusive, but each corresponds to a different computational assumption. \emph{Rate coding} assumes information is represented by firing rate: for a spike train \(s(t)=\sum_f \delta(t-t^{(f)})\), the rate over a window of length \(\Delta\) is \(r_\Delta(t) = \Delta^{-1}\int_{t-\Delta}^{t} s(\tau)\,d\tau\) \cite{adrian1926impulses}, and the scalar activation of an artificial unit can be read as a rate-like abstraction. \emph{Temporal coding} assumes information is carried also by when spikes occur; cortical neurons can produce reliable spike timing across repeated presentations \cite{mainen1995reliability}, with information in first-spike latency, synchrony, and interspike intervals \cite{panzeri2010sensory}, and this scheme motivates \glspl{snn} \cite{maass1997networks}. \emph{Population coding} assumes information is represented collectively by neurons with broad overlapping tuning \cite{georgopoulos1986neuronal} and is implicit whenever a model represents state as a high-dimensional vector.
 
The coding scheme constrains which learning rules can be defined locally: Hebbian plasticity treats co-activity as the primitive relation, and \gls{stdp} refines this by making synaptic change depend on relative spike timing \cite{hebb1949organization,bi1998synaptic,dan2004spike}; these rules are meaningful only relative to the temporal structure retained by the forward model.
 
\subsection{Neural Computation as State Evolution}
\label{sec:foundations_state}
 
A biological or artificial neural system can be described as a system whose internal state \(\mathbf{h}(t)\) evolves under an input-driven rule:
\begin{equation}
\dot{\mathbf{h}}(t) = F\bigl(\mathbf{h}(t),\, \mathbf{x}(t);\, \theta_F\bigr),
\qquad
\mathbf{y}(t) = G\bigl(\mathbf{h}(t);\, \theta_G\bigr),
\label{eq:state_evolution_continuous}
\end{equation}
with \(\mathbf{x}(t)\) the input, \(\mathbf{h}(t)\) the hidden state, \(\mathbf{y}(t)\) the output, \(F\) the dynamics, \(G\) the output map, and \(\theta_F,\theta_G\) parameters. In a \gls{lif} neuron, \(\mathbf{h}(t)\) is the membrane potential \(V(t)\), \(F\) is the right-hand side of Eq.~\eqref{eq:lif_single}, and the output is a spike emitted at threshold crossing. In a continuous-time artificial model such as a Neural ODE, \(\mathbf{h}(t)\) is a hidden vector and \(F\) is a learned vector field.
 
Discrete-time recurrent models have the analogous form
\begin{equation}
\mathbf{h}_{t+1} = F\bigl(\mathbf{h}_t,\, \mathbf{x}_t;\, \theta_F\bigr),
\qquad
\mathbf{y}_t = G\bigl(\mathbf{h}_t;\, \theta_G\bigr).
\label{eq:state_evolution_discrete}
\end{equation}
A static feedforward network is the degenerate case with no persistent state, reducing to the memoryless map \(\mathbf{y} = f_\theta(\mathbf{x})\) of Eq.~\eqref{eq:intro_static_mapping}. Hybrid event-driven systems combine continuous evolution with discrete jumps or resets. Implicit-state models define their latent representation as the solution of a fixed-point equation
\begin{equation}
\mathbf{z}^{\star} = F(\mathbf{z}^{\star}, \mathbf{x}; \theta_F),
\label{eq:implicit_state}
\end{equation}
rather than as the result of explicit time stepping.
 
The state-dynamics axis of Section~\ref{sec:taxonomy} distinguishes static, discrete-time, continuous-time, implicit, and hybrid event-driven computation; the learning axis tracks the credit-assignment mechanism that attributes changes in an objective, reward, or prediction error to internal variables and parameters. These axes are coupled: the choice of dynamics determines which gradients, adjoints, surrogate derivatives, or local traces are available, while the available learning rule constrains which dynamics can be trained at scale. This coupling is the technical basis of the forward--backward disconnect.
 
\subsection{Joint Scope: Dynamics and Biological Grounding}
\label{sec:foundations_joint}
 
The scope statement of Section~\ref{sec:introduction} admits a model family if it instantiates a nontrivial form of state dynamics or is substantively grounded in neuroscience. These criteria are treated jointly because, in biological systems, dynamics and mechanism are inseparable: membrane potentials evolve according to differential equations because ion channels and synapses have physical time constants; spike-based communication is event-driven because action potentials are threshold-triggered nonlinear events; dendritic processing is state-dependent because signals propagate through spatially extended active compartments. Surveys organized only around biological inspiration underemphasize engineering-driven dynamical architectures such as structured \glspl{ssm}, while surveys organized only around dynamical systems underemphasize predictive coding, Hierarchical Temporal Memory, and local plasticity; the joint view places these families in a common space and treats biological grounding both as a scope condition and as a diagnostic that distinguishes mathematical choices with biological consequences from those that are merely convenient abstractions.
 
The scope is fixed by an architecture-independent inclusion rule: a model family is admitted if it satisfies at least one of (i) its internal computation instantiates a nontrivial state-evolution mechanism central to the architecture, in the sense of Section~\ref{sec:foundations_state}, or (ii) a biological mechanism materially constrains its architecture or learning rule, in the sense of the forward or learning grounding tiers of Section~\ref{sec:taxonomy_bio}. \glspl{pinn} are the principal borderline case: the network is a static coordinate-to-field map, but the dynamics are encoded into the training objective (Eq.~\eqref{eq:static_pinn_loss}), a mechanism that recurs with energy-based objectives in Section~\ref{sec:learning}. Each model-family section that follows asks three biology-as-discussant questions: what computational problem does biology solve, which features of that solution does the artificial family preserve, and which does it omit and at what compensating cost.

\section{Taxonomy and Related Survey Positioning}
\label{sec:taxonomy}
 
Section~\ref{sec:foundations} established the dynamical-systems vocabulary in which neural models can be uniformly described as state-evolution systems. This section turns that vocabulary into a taxonomy, populates it with the model families covered by the rest of the survey, and positions the present work against existing reviews of overlapping territory.
 
\subsection{Taxonomy by State-Dynamics Structure}
\label{sec:taxonomy_dynamics}
 
The horizontal axis of the taxonomy partitions neural models by the form of their internal state evolution, following directly from the state-evolution descriptions in Eqs.~\eqref{eq:state_evolution_continuous}--\eqref{eq:implicit_state}. Five classes are distinguished.
 
\emph{Static} models implement memoryless input--output maps \(\mathbf{y} = f_\theta(\mathbf{x})\) with no persistent internal state across evaluations, recovering Eq.~\eqref{eq:intro_static_mapping}. \emph{Discrete-time / sequence} models operate over a discrete time or position index, either through explicit recurrence with a hidden state \(\mathbf{h}_t\) updated by a difference equation (RNN, LSTM, GRU) or through a sequence-level computation that admits a recurrent interpretation (Transformer at inference, structured state-space models at training and inference). \emph{Continuous-time} models describe the state \(\mathbf{h}(t)\) as a trajectory governed by differential equations. \emph{Implicit-state} models define the latent representation \(\mathbf{z}^\star\) as the solution of a fixed-point equation. \emph{Hybrid event-driven} models combine continuous subthreshold state evolution with discrete events such as spikes or resets.
 
These five classes are not arbitrary partitions of a continuum. Each corresponds to a qualitatively distinct mathematical object and places different demands on the credit-assignment mechanism.
 
\subsection{Learning-Rule and Credit-Assignment Axis}
\label{sec:taxonomy_learning}
 
The vertical axis tracks the mechanism by which model parameters are adjusted given training signals. Four categories are sufficient for the model families covered in this survey.
 
\emph{Global gradient} mechanisms use backpropagation \cite{rumelhart1986learning} or its temporal extension \gls{bptt} to compute exact gradients of a global objective. \emph{Approximate or implicit gradient} mechanisms cover two related departures: \emph{approximate} variants that compute an inexact but useful gradient signal (feedback alignment \cite{lillicrap2016random}, target propagation, and surrogate gradients used in spiking networks), and \emph{implicit} variants that recover gradients through non-standard machinery, exact under their defining conditions or limits rather than through layered backpropagation (equilibrium propagation \cite{scellier2017equilibrium} and adjoint methods used in continuous-time models). What unites the row is that none of these methods executes a standard forward--backward graph traversal with exact local derivatives; each replaces that operation with either an approximation or an alternative exact-under-conditions procedure. \emph{Local plasticity} rules update synaptic weights from variables available at the synapse: Hebbian learning \cite{hebb1949organization}, spike-timing-dependent plasticity \cite{bi1998synaptic,dan2004spike}, and dendrite-local rules. \emph{Energy/unsupervised} objectives define learning through energy minimization or local goodness functions: contrastive divergence in restricted Boltzmann machines, the unsupervised loss in Forward-Forward \cite{hinton2022forward}.
 
The four-row primary is a compact projection of a multi-dimensional structure. Five secondary attributes separate rows that the primary label conflates: \emph{gradient relation} (exact / approximate / non-gradient), \emph{update scope} (synapse-local vs.\ requiring information about other weights or activations), \emph{temporal requirement} (whether the rule presupposes temporal structure such as spike timing, two-phase relaxation, or unrolled sequences), \emph{objective source} (externally supervised, internally derived from a goodness or energy function, or unsupervised), and \emph{update timing} (online, between two forward passes, or once after a full forward--backward traversal). The four-row primary is retained for compactness of exposition and Figure~\ref{fig:master_taxonomy}; the secondary attributes make the heterogeneity explicit without requiring the compact primary to encode all dimensions simultaneously, and Supplementary Section~S3 reports the per-configuration ledger that generates Figure~\ref{fig:master_taxonomy}'s counts. Table~\ref{tab:secondary_dims} instantiates these five secondary dimensions for representative mechanisms, so a reader can locate any new learning rule along the same coordinates rather than into the four-row primary alone.
 
\begin{table*}[!htbp]
\caption{Secondary credit-assignment dimensions applied to representative learning mechanisms. These attributes are descriptive coordinates rather than additional mutually exclusive classes; individual variants may differ. The table instantiates the multidimensional structure of the backward axis introduced above, so a new architecture--learning configuration can be located along the same coordinates.}
\label{tab:secondary_dims}
\centering
\scriptsize
\setlength{\tabcolsep}{2pt}
\renewcommand{\arraystretch}{1.10}
\begin{tabular}{@{}>{\raggedright\arraybackslash}p{2.3cm} >{\raggedright\arraybackslash}p{1.9cm} >{\raggedright\arraybackslash}p{2.2cm} >{\raggedright\arraybackslash}p{2.1cm} >{\raggedright\arraybackslash}p{1.9cm} >{\raggedright\arraybackslash}p{2.2cm}@{}}
\toprule
\textbf{Mechanism} & \textbf{Gradient relation} & \textbf{Update scope} & \textbf{Temporal requirement} & \textbf{Objective source} & \textbf{Update timing} \\
\midrule
Backpropagation \cite{rumelhart1986learning} & Exact & Global / non-local & None (static graph) & Global supervised & After forward pass \\
\gls{bptt} \cite{werbos1990backpropagation} & Exact & Global / non-local & Unrolled sequence & Global supervised & After full sequence \\
FA / DFA / PFA \cite{lillicrap2016random,nokland2016direct,li2024deep} & Approximate & Broadcast error via random or learned matrices & None in feedforward case & Global supervised & Feedback pass; DFA partly parallel \\
Synthetic gradients \cite{jaderberg2017decoupled} & Approximate (predicted) & Module-local predictor; target derived from downstream gradient & Optional recurrent state & Global supervised & Asynchronous \\
Adjoint / implicit diff. \cite{chen2018neural,bai2019deep} & Exact under solver / equilibrium conditions & Solver / global linear solve at equilibrium & Trajectory or fixed point & Global objective & Backward solve after forward solve \\
Surrogate-gradient \gls{snn} \cite{neftci2019surrogate} & Approximate derivative & Global error propagation through smoothed spikes & Spike-time trajectory & Global supervised & Backward through temporal graph \\
Equilibrium propagation \cite{scellier2017equilibrium} / predictive coding \cite{whittington2017approximation} & Exact only in specified limits; approximate otherwise & Local update after network relaxation & Two-phase / iterative inference & Supervised + energy/error & Between/after relaxation phases \\
Hebbian / \gls{stdp} \cite{hebb1949organization,bi1998synaptic} & Non-gradient & Synapse-local & Co-activity or relative spike timing & Local or unsupervised & Online \\
Three-factor / e-prop \cite{fremaux2016neuromodulated,bellec2020solution} & Variant-dependent: non-gradient to gradient-inspired approximate & Local eligibility trace $+$ modulatory or broadcast signal & Eligibility traces over time & Supervised or reward-modulated & Online \\
Forward-Forward \cite{hinton2022forward} & Local-objective gradient & Layer-local goodness function & Two forward phases (positive / negative) & Local goodness objective & Between forward passes \\
\bottomrule
\end{tabular}
\end{table*}
 
Within each of these four categories, the credit-assignment problem itself has two distinguishable sub-problems. \emph{Spatial credit assignment} attributes the change in a loss to parameters across the depth or width of the forward graph at a single time. \emph{Temporal credit assignment} attributes the change in a loss to parameters across time, where the same parameters affect the output at many later steps.
 
Static feedforward models face only the spatial problem. Discrete-time recurrent and continuous-time models face both, but \gls{bptt} reduces temporal credit assignment to spatial credit assignment by unrolling time into a layered graph with tied weights, and this reduction is exactly where the biological implausibility peaks: biological systems do not appear to retain and replay a complete ordered computational graph of prior activations for exact reverse-mode differentiation in the manner required by \gls{bptt}. Hybrid event-driven spiking models face both problems simultaneously, with the additional complication of non-differentiable spike events. The biologically plausible learning rules surveyed in Section~\ref{sec:learning} are best read as alternative solutions to one or both of these sub-problems: in particular, alternatives that do not unroll time into space.
 
The two axes are coupled. The form of the forward dynamics restricts the gradient operators that are available: continuous-time models admit adjoint methods, fixed-point models admit implicit differentiation, hybrid spiking models require surrogates or local rules. Conversely, the learning rule restricts which dynamics can be trained at scale.
 
Figure~\ref{fig:master_taxonomy} is organized as a bipartite flow in which each \emph{combination} of a state-dynamics class (source) and a credit-assignment mechanism (target) defines one position in the joint taxonomy, and the same forward architecture can occupy different combinations depending only on how it is trained. A spiking network trained with surrogate gradients and the same spiking architecture trained with \gls{stdp} enter the \emph{Approximate or implicit gradient} and \emph{Local plasticity} targets respectively; an \gls{mlp} trained by backpropagation and by Forward-Forward likewise enter different targets. Treating such cases as identical entries would conceal the credit-assignment asymmetry the taxonomy is designed to expose.
 
The atomic unit of classification in this survey is therefore the \emph{architecture--learning configuration} rather than the architecture name in isolation. A model family appears in more than one taxonomy combination when distinct training mechanisms substantively alter its credit-assignment behavior, and a model-agnostic learning rule (feedback alignment, Forward-Forward, equilibrium propagation, predictive coding) is placed in the taxonomy only after specifying the forward architecture on which it is evaluated. The same commitment extends to the biological-grounding axis introduced in Section~\ref{sec:taxonomy_bio}, where forward grounding and learning grounding are assessed separately for each configuration. Placement decisions for ambiguous configurations (modern Hopfield, predictive-coding variants, equilibrium propagation, Forward-Forward, the surrogate-gradient and \gls{stdp}-trained spiking variants) are documented with their rationale in the supplementary material.
 
\subsection{Biological Grounding as a Categorical Axis}
\label{sec:taxonomy_bio}
 
Biological grounding is treated categorically and, per the architecture--learning configuration unit of Section~\ref{sec:taxonomy_learning}, applied independently to the forward operator and to the credit-assignment rule. The two axes share an ordinal structure of four tiers each, in increasing order of biological constraint.
 
The forward-grounding axis classifies the architecture of the forward computation:
 
\begin{itemize}[leftmargin=*, itemsep=0pt, topsep=2pt]
\item \emph{None or historical inspiration}: biology either did not motivate the model or motivated it historically without constraining the modern design. Examples: structured \glspl{ssm}, \glspl{deq}, the \gls{mlp}, the modern \gls{cnn}.
\item \emph{Functional analogy}: the model resembles a biological process at a functional level but the resemblance is post-hoc, and the forward operator would have been adopted independently for engineering reasons. Examples: \glspl{rnn} as analogue to recurrent cortical circuits; modern Hopfield as attractor analogue to attention.
\item \emph{Architectural constraint}: a specific biological structure from Section~\ref{sec:foundations_neurons} or~\ref{sec:foundations_coding} constrains the architecture in a way the design would not otherwise adopt. Examples: the Neocognitron's simple/complex cell alternation; classical Hopfield networks.
\item \emph{Mechanistic plausibility}: the unit-level computation is tied to biophysical mechanisms (membrane-potential dynamics, threshold-triggered spikes, dendritic compartments). Example: \gls{lif}-based spiking neural networks.
\end{itemize}
 
The learning-grounding axis classifies the credit-assignment rule:
 
\begin{itemize}[leftmargin=*, itemsep=0pt, topsep=2pt]
\item \emph{None}: no biological motivation. Examples: backpropagation, \gls{bptt}.
\item \emph{Weak}: loose biological motivation without locality or empirical-plasticity grounding. Example: feedback alignment.
\item \emph{Local or algorithmic}: the rule is local in synaptic weights or biologically motivated through co-activity but is not derived from empirical plasticity measurements. Examples: contrastive divergence, equilibrium propagation, predictive-coding variants, Forward-Forward (layerwise-local goodness objective), Hebbian storage in classical Hopfield networks, dendrite-local Hebbian permanence updates in HTM.
\item \emph{Mechanistic plasticity}: the rule is directly grounded in measured synaptic plasticity. Example: \gls{stdp}.
\end{itemize}
 
The principal tier-boundary distinctions: on the forward axis, \emph{Architectural constraint} requires biology to rule out alternative design choices (the Neocognitron's simple/complex cell alternation), distinguishing it from \emph{Functional analogy} (the \gls{rnn}'s recurrence, which would have been adopted independently for sequence modeling). On the learning axis, \emph{Mechanistic plasticity} requires the rule itself to be empirically observed plasticity rather than a locally-consistent algorithmic substitute.
 
Table~\ref{tab:model_families} summarizes the principal model families and their grounding assignments; the complete configuration-level ledger is in Supplementary Section~S3.
 
Figure~\ref{fig:master_taxonomy} presents the joint taxonomy as an alluvial flow between forward state-dynamics classes (left) and credit-assignment mechanisms (right), with node heights and ribbon widths proportional to configuration counts. Table~\ref{tab:model_families} summarizes the principal families; Supplementary Section~S3 gives the full configuration-level ledger. The forward side distributes across five classes with node heights ranging from 2 to 14, and every forward class sends at least one ribbon somewhere on the right. The right side has a different shape: Global gradient is the largest single credit-assignment category, accounting for half of the audited configurations, receives ribbons from four of the five forward classes, and is followed by Approximate or implicit gradient as the second-largest category. This many-to-one bottleneck between the forward fan and the backward funnel is the forward--backward disconnect introduced in Section~\ref{sec:introduction} and developed further in Section~\ref{sec:learning}.
 
\begin{figure*}[t]
\centering
\begin{adjustbox}{max width=\textwidth}
\begin{tikzpicture}[
  x=1cm, y=1cm,
  flabel/.style = {anchor=east, font=\small\bfseries, align=right},
  fcount/.style = {anchor=east, font=\scriptsize, text=black!55},
  blabel/.style = {anchor=west, font=\small\bfseries, align=left},
  bcount/.style = {anchor=west, font=\scriptsize, text=black!55},
  colhead/.style = {anchor=south, font=\small\bfseries\itshape, text=black!70}
]
 
\def\lx{0}       
\def\lw{0.45}    
\def\rx{10}      
\def\rw{0.45}    
\def\opac{0.42}   
\def\cbA{3.5}    
 
 
\node[colhead] at (\lx+\lw/2, 0.7) {State dynamics (forward)};
\node[colhead] at (\rx+\rw/2, 0.7) {Credit assignment (backward)};
 
\fill[cStatic] (\lx, 0) rectangle (\lx+\lw, -1.5);
\draw[black!70, line width=0.4pt] (\lx, 0) rectangle (\lx+\lw, -1.5);
\node[flabel] at (\lx-0.15, -0.65) {Static};
\node[fcount] at (\lx-0.15, -0.95) {6 configs.};
 
\fill[cDiscrete] (\lx, -1.75) rectangle (\lx+\lw, -5.25);
\draw[black!70, line width=0.4pt] (\lx, -1.75) rectangle (\lx+\lw, -5.25);
\node[flabel, align=right] at (\lx-0.15, -3.3) {Discrete-time /\\sequence};
\node[fcount] at (\lx-0.15, -3.9) {14 configs.};
 
\fill[cContinuous] (\lx, -5.5) rectangle (\lx+\lw, -7.0);
\draw[black!70, line width=0.4pt] (\lx, -5.5) rectangle (\lx+\lw, -7.0);
\node[flabel] at (\lx-0.15, -6.15) {Continuous-time};
\node[fcount] at (\lx-0.15, -6.45) {6 configs.};
 
\fill[cImplicit] (\lx, -7.25) rectangle (\lx+\lw, -8.25);
\draw[black!70, line width=0.4pt] (\lx, -7.25) rectangle (\lx+\lw, -8.25);
\node[flabel] at (\lx-0.15, -7.65) {Implicit (fixed-point)};
\node[fcount] at (\lx-0.15, -7.95) {4 configs.};
 
\fill[cHybrid] (\lx, -8.5) rectangle (\lx+\lw, -9.0);
\draw[black!70, line width=0.4pt] (\lx, -8.5) rectangle (\lx+\lw, -9.0);
\node[flabel] at (\lx-0.15, -8.65) {Hybrid event-driven};
\node[fcount] at (\lx-0.15, -8.9) {2 configs.};
 
\fill[black!40] (\rx, -0.125) rectangle (\rx+\rw, -4.125);
\draw[black!70, line width=0.4pt] (\rx, -0.125) rectangle (\rx+\rw, -4.125);
\node[blabel] at (\rx+\rw+0.15, -2.0) {Global gradient};
\node[bcount] at (\rx+\rw+0.15, -2.35) {16 configs.};
 
\fill[black!40] (\rx, -4.375) rectangle (\rx+\rw, -6.125);
\draw[black!70, line width=0.4pt] (\rx, -4.375) rectangle (\rx+\rw, -6.125);
\node[blabel, align=left] at (\rx+\rw+0.15, -5.05) {Approximate or\\implicit gradient};
\node[bcount] at (\rx+\rw+0.15, -5.75) {7 configs.};
 
\fill[black!40] (\rx, -6.375) rectangle (\rx+\rw, -7.875);
\draw[black!70, line width=0.4pt] (\rx, -6.375) rectangle (\rx+\rw, -7.875);
\node[blabel] at (\rx+\rw+0.15, -7.0) {Local plasticity};
\node[bcount] at (\rx+\rw+0.15, -7.35) {6 configs.};
 
\fill[black!40] (\rx, -8.125) rectangle (\rx+\rw, -8.875);
\draw[black!70, line width=0.4pt] (\rx, -8.125) rectangle (\rx+\rw, -8.875);
\node[blabel] at (\rx+\rw+0.15, -8.4) {Energy / unsupervised};
\node[bcount] at (\rx+\rw+0.15, -8.7) {3 configs.};
 
\fill[cStatic, opacity=\opac]
  (\lx+\lw, 0)
  .. controls (\lx+\lw+\cbA, 0) and (\rx-\cbA, -0.125) .. (\rx, -0.125)
  -- (\rx, -1.125)
  .. controls (\rx-\cbA, -1.125) and (\lx+\lw+\cbA, -1.0) .. (\lx+\lw, -1.0)
  -- cycle;
 
\fill[cStatic, opacity=\opac]
  (\lx+\lw, -1.0)
  .. controls (\lx+\lw+\cbA, -1.0) and (\rx-\cbA, -6.375) .. (\rx, -6.375)
  -- (\rx, -6.625)
  .. controls (\rx-\cbA, -6.625) and (\lx+\lw+\cbA, -1.25) .. (\lx+\lw, -1.25)
  -- cycle;
 
\fill[cStatic, opacity=\opac]
  (\lx+\lw, -1.25)
  .. controls (\lx+\lw+\cbA, -1.25) and (\rx-\cbA, -8.125) .. (\rx, -8.125)
  -- (\rx, -8.375)
  .. controls (\rx-\cbA, -8.375) and (\lx+\lw+\cbA, -1.5) .. (\lx+\lw, -1.5)
  -- cycle;
 
\fill[cDiscrete, opacity=\opac]
  (\lx+\lw, -1.75)
  .. controls (\lx+\lw+\cbA, -1.75) and (\rx-\cbA, -1.125) .. (\rx, -1.125)
  -- (\rx, -3.125)
  .. controls (\rx-\cbA, -3.125) and (\lx+\lw+\cbA, -3.75) .. (\lx+\lw, -3.75)
  -- cycle;
 
\fill[cDiscrete, opacity=\opac]
  (\lx+\lw, -3.75)
  .. controls (\lx+\lw+\cbA, -3.75) and (\rx-\cbA, -4.375) .. (\rx, -4.375)
  -- (\rx, -4.875)
  .. controls (\rx-\cbA, -4.875) and (\lx+\lw+\cbA, -4.25) .. (\lx+\lw, -4.25)
  -- cycle;
 
\fill[cDiscrete, opacity=\opac]
  (\lx+\lw, -4.25)
  .. controls (\lx+\lw+\cbA, -4.25) and (\rx-\cbA, -6.625) .. (\rx, -6.625)
  -- (\rx, -7.125)
  .. controls (\rx-\cbA, -7.125) and (\lx+\lw+\cbA, -4.75) .. (\lx+\lw, -4.75)
  -- cycle;
 
\fill[cDiscrete, opacity=\opac]
  (\lx+\lw, -4.75)
  .. controls (\lx+\lw+\cbA, -4.75) and (\rx-\cbA, -8.375) .. (\rx, -8.375)
  -- (\rx, -8.875)
  .. controls (\rx-\cbA, -8.875) and (\lx+\lw+\cbA, -5.25) .. (\lx+\lw, -5.25)
  -- cycle;
 
\fill[cContinuous, opacity=\opac]
  (\lx+\lw, -5.5)
  .. controls (\lx+\lw+\cbA, -5.5) and (\rx-\cbA, -3.125) .. (\rx, -3.125)
  -- (\rx, -3.875)
  .. controls (\rx-\cbA, -3.875) and (\lx+\lw+\cbA, -6.25) .. (\lx+\lw, -6.25)
  -- cycle;
 
\fill[cContinuous, opacity=\opac]
  (\lx+\lw, -6.25)
  .. controls (\lx+\lw+\cbA, -6.25) and (\rx-\cbA, -4.875) .. (\rx, -4.875)
  -- (\rx, -5.125)
  .. controls (\rx-\cbA, -5.125) and (\lx+\lw+\cbA, -6.5) .. (\lx+\lw, -6.5)
  -- cycle;
 
\fill[cContinuous, opacity=\opac]
  (\lx+\lw, -6.5)
  .. controls (\lx+\lw+\cbA, -6.5) and (\rx-\cbA, -7.125) .. (\rx, -7.125)
  -- (\rx, -7.625)
  .. controls (\rx-\cbA, -7.625) and (\lx+\lw+\cbA, -7.0) .. (\lx+\lw, -7.0)
  -- cycle;
 
\fill[cImplicit, opacity=\opac]
  (\lx+\lw, -7.25)
  .. controls (\lx+\lw+\cbA, -7.25) and (\rx-\cbA, -3.875) .. (\rx, -3.875)
  -- (\rx, -4.125)
  .. controls (\rx-\cbA, -4.125) and (\lx+\lw+\cbA, -7.5) .. (\lx+\lw, -7.5)
  -- cycle;
 
\fill[cImplicit, opacity=\opac]
  (\lx+\lw, -7.5)
  .. controls (\lx+\lw+\cbA, -7.5) and (\rx-\cbA, -5.125) .. (\rx, -5.125)
  -- (\rx, -5.875)
  .. controls (\rx-\cbA, -5.875) and (\lx+\lw+\cbA, -8.25) .. (\lx+\lw, -8.25)
  -- cycle;
 
\fill[cHybrid, opacity=\opac]
  (\lx+\lw, -8.5)
  .. controls (\lx+\lw+\cbA, -8.5) and (\rx-\cbA, -5.875) .. (\rx, -5.875)
  -- (\rx, -6.125)
  .. controls (\rx-\cbA, -6.125) and (\lx+\lw+\cbA, -8.75) .. (\lx+\lw, -8.75)
  -- cycle;
 
\fill[cHybrid, opacity=\opac]
  (\lx+\lw, -8.75)
  .. controls (\lx+\lw+\cbA, -8.75) and (\rx-\cbA, -7.625) .. (\rx, -7.625)
  -- (\rx, -7.875)
  .. controls (\rx-\cbA, -7.875) and (\lx+\lw+\cbA, -9.0) .. (\lx+\lw, -9.0)
  -- cycle;
 
\useasboundingbox (-4, 1.1) rectangle (14, -9.3);
 
\end{tikzpicture}
\end{adjustbox}
\caption{Joint taxonomy of neural computation as an alluvial flow. Each architecture--learning configuration is one unit of flow from a forward state-dynamics class (left column) to a credit-assignment mechanism (right column). Node heights and ribbon widths are proportional to configuration counts; ribbon colors identify the source class. Table~\ref{tab:model_families} lists model families and grounding assignments; Supplementary Section~S3 provides the complete configuration-level ledger. Global gradient is the largest single credit-assignment category (16/32), followed by Approximate or implicit gradient (7/32), and it receives configurations from four of the five forward classes: the categorical form of the forward--backward disconnect introduced in Section~\ref{sec:introduction}.}
\Description{An alluvial flow diagram showing the joint taxonomy as a bipartite flow. Five colored vertical bars on the left represent forward state-dynamics classes, from top to bottom: Static (6 configurations, blue), Discrete-time or sequence (14, orange), Continuous-time (6, green), Implicit fixed-point (4, purple), Hybrid event-driven (2, red). Four gray vertical bars on the right represent credit-assignment mechanisms, from top to bottom: Global gradient (16 configurations), Approximate or implicit gradient (7), Local plasticity (6), Energy or unsupervised (3). Fourteen curved ribbons connect the left classes to the right mechanisms; each ribbon's width is proportional to how many configurations flow from that source to that destination, and each ribbon is colored by its source class. Global gradient is the largest single credit-assignment category, accounting for half of the audited configurations and receiving ribbons from four of the five forward classes, followed by Approximate or implicit gradient as the second-largest. The visual impression is a fan of colored streams on the left converging predominantly into the two upper destinations on the right, which is the visual signature of the credit-assignment concentration relative to forward-dynamics diversification.}
\label{fig:master_taxonomy}
\end{figure*}
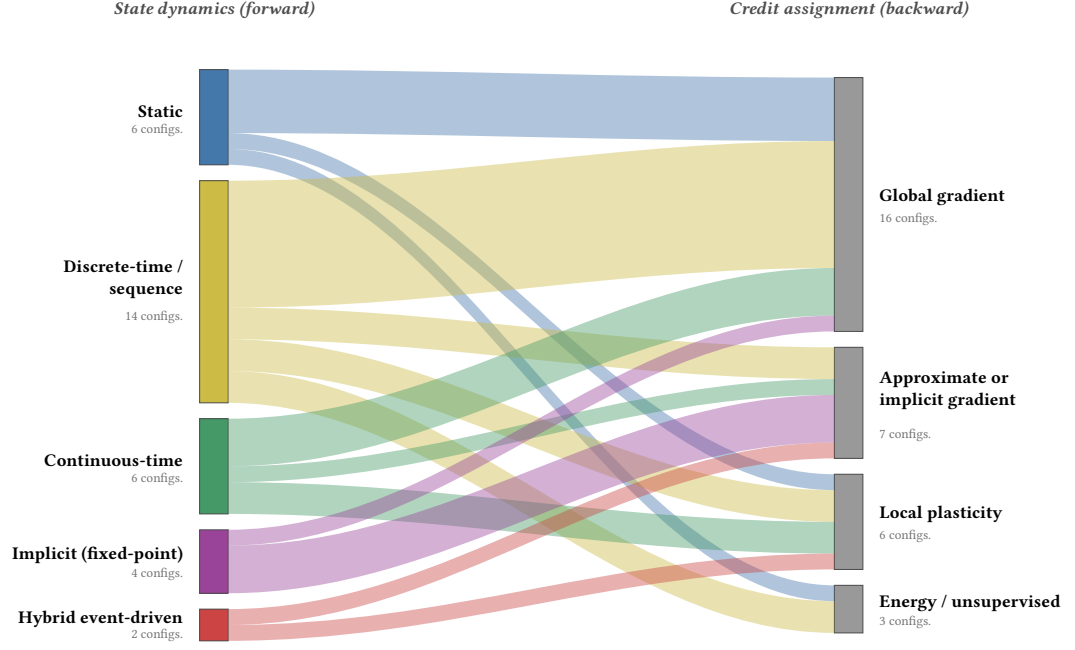
 
\begin{table*}[!htbp]
\caption{Configuration counts underlying Figure~\ref{fig:master_taxonomy}, with the model families that populate each cell. Cell entries follow the pattern \emph{count: family list (C-IDs referencing the per-configuration ledger of Supplementary Section~S3)}. Row and column totals match the node heights of Figure~\ref{fig:master_taxonomy} exactly.}
\label{tab:config_count_matrix}
\centering
\scriptsize
\setlength{\tabcolsep}{2pt}
\renewcommand{\arraystretch}{1.05}
\begin{tabular}{@{}>{\raggedright\arraybackslash}p{1.3cm} >{\raggedright\arraybackslash}p{3.7cm} >{\raggedright\arraybackslash}p{3.0cm} >{\raggedright\arraybackslash}p{2.3cm} >{\raggedright\arraybackslash}p{2.2cm} c@{}}
\toprule
\textbf{Forward class} & \textbf{Global gradient} & \makecell[l]{\textbf{Approximate or}\\\textbf{implicit gradient}} & \textbf{Local plasticity} & \makecell[l]{\textbf{Energy /}\\\textbf{unsupervised}} & \textbf{Total} \\
\midrule
Static      & 4: MLP, CNN, ResNet, PINN (C01--C04) & 0 & 1: Neocognitron (C05) & 1: MLP + Forward-Forward (C06) & 6 \\
Discrete    & 8: RNN, LSTM, GRU, Transformer, S4, Mamba, RWKV, xLSTM (C07--C14) & 2: RNN + synth.\ grad., RNN + RFLO (C15, C16) & 2: classical Hopfield, HTM (C17, C32) & 2: RBM, DBN (C18, C19) & 14 \\
Continuous  & 3: CTRNN, Liquid TC, Liquid closed-form (C20--C22) & 1: Neural ODE (C23) & 2: PC recurrent, CT-RNN + reward-mod.\ Hebb.\ (C24, C25) & 0 & 6 \\
Implicit    & 1: modern Hopfield (C26) & 3: DEQ, PC local-error, PC BP-equiv.\ limit (C27--C29) & 0 & 0 & 4 \\
Hybrid      & 0 & 1: surrogate-gradient LIF SNN (C30) & 1: STDP-trained LIF SNN (C31) & 0 & 2 \\
\midrule
\textbf{Total} & \textbf{16} & \textbf{7} & \textbf{6} & \textbf{3} & \textbf{32} \\
\bottomrule
\end{tabular}
\end{table*}
 
A finer mechanism-level decomposition reported in Supplementary Section~S3 preserves the qualitative asymmetry: the Approximate or implicit gradient row fragments across several individually small mechanisms (synthetic gradients, random-feedback local online learning, adjoint sensitivity, implicit differentiation, predictive-coding approximations, and surrogate gradients), whereas ordinary reverse-mode backpropagation remains the largest single mechanism in the survey.
 
\begin{table}[p]
\caption{Model families covered in this survey. Column definitions follow Sections~\ref{sec:taxonomy_dynamics}--\ref{sec:taxonomy_bio}. Biological grounding is split into a forward axis and a learning axis, applied independently to each architecture--learning configuration. Configurations whose placement required adjudication (modern Hopfield, predictive-coding variants, equilibrium propagation, Forward-Forward, surrogate-gradient and \gls{stdp}-trained spiking variants) are documented in the supplementary material.}
\label{tab:model_families}
\centering
\scriptsize
\setlength{\tabcolsep}{4pt}
\renewcommand{\arraystretch}{1.15}
\begin{adjustbox}{angle=90, max totalheight=0.88\textheight, max width=\textwidth, center}
\begin{tabular}{@{}p{2.4cm}p{3.0cm}p{2.0cm}p{2.3cm}p{2.3cm}p{2.5cm}p{1.7cm}p{1.7cm}p{2.5cm}@{}}
\toprule
\textbf{Family} & \textbf{Representative models} & \textbf{Time repr.} & \textbf{State variable} & \textbf{Communication} & \textbf{Learning rule} & \textbf{Forward gnd.} & \textbf{Learning gnd.} & \textbf{Scalability bottleneck} \\
\midrule
Static feedforward & MLP, CNN, ResNet \cite{rumelhart1986learning,lecun1998gradient,he2016deep} & Static & None & Dense real-valued & BP & None/Hist. & None & Memory cost of activations \\
Static, bio-inspired & Neocognitron \cite{fukushima1980neocognitron} & Static & None & Dense real-valued & Local Hebbian & Arch.\ constr. & Local/alg. & Depth, no global gradient \\
Static, physics-constrained & PINN \cite{raissi2019physics} & Static & None & Dense real-valued & BP + PDE residual & None/Hist. & None & PDE residual conditioning \\
Discrete recurrent & RNN, LSTM, GRU \cite{elman1990finding,hochreiter1997long,cho2014learning} & Discrete & Hidden state & Dense real-valued & BPTT & Funct.\ analogy & None & Vanishing gradients, BPTT \\
Classical Hopfield & Hopfield 1982 \cite{hopfield1982neural} & Discrete (fixed-pt) & State vector & Dense real-valued & Hebbian storage & Arch.\ constr. & Local/alg. & Capacity $\approx 0.14 N$ \cite{amit1985storing} \\
Modern Hopfield & Ramsauer 2021 \cite{ramsauer2021hopfield} & Implicit / fixed-pt & State vector & Dense real-valued & End-to-end BP & Funct.\ analogy & None & Attention-like retrieval cost \\
Attention-based & Transformer \cite{vaswani2017attention} & Discrete (pos.) & Key--value cache & Dense real-valued & BP & None/Hist. & None & Quadratic attn.\ cost \\
State-space sequence & S4, Mamba, RWKV, xLSTM \cite{gu2022efficiently,gu2023mamba,peng2023rwkv,beck2024xlstm} & Discrete; cont.-time param. & Latent SSM state & Dense real-valued & BP & None/Hist. & None & State-update parallelism \\
Continuous-time recurrent & CTRNN, Liquid NN \cite{funahashi1993approximation,hasani2021liquid} & Continuous & Cont.\ trajectory & Dense real-valued & Solver BP / BPTT & Funct.\ analogy & None & Solver cost, stiffness \\
Continuous-depth & Neural ODE \cite{chen2018neural} & Continuous & Cont.\ trajectory & Dense real-valued & Adjoint / solver BP & None/Hist. & None & Solver cost, stiffness \\
Reservoir / liquid-state & ESN, LSM \cite{jaeger2001echo,maass2002real} & Discrete (ESN) / hybrid event-driven (LSM) & Reservoir state & Cont.\ or spike & Readout-only & Funct.\ analogy & None & Fixed-reservoir expressivity \\
Fixed-point / implicit & DEQ \cite{bai2019deep} & Implicit & Equilibrium state & Dense real-valued & Implicit diff. & None/Hist. & None & Solver convergence \\
Surrogate-grad SNN & LIF networks \cite{maass1997networks,neftci2019surrogate} & Hybrid event-driv. & Membrane potential & Sparse binary spikes & Surrogate gradient & Mech.\ plaus. & Weak & Non-diff.\ spikes \\
STDP-trained SNN & LIF + STDP \cite{bi1998synaptic,markram1997regulation} & Hybrid event-driv. & Membrane potential & Sparse binary spikes & Spike-timing-dep. & Mech.\ plaus. & Mech.\ plast. & Scaling at depth \\
Predictive coding / AIF & PC, FEP \cite{rao1999predictive,friston2010free,whittington2017approximation,millidge2022predictive} & Discrete / cont. & Local error + state & Dense real-valued & Local error min. & Arch.\ / Funct. & Local/alg. & Convergence, scaling \\
Hierarchical Temporal Memory & HTM, Thousand Brains \cite{hawkins2016neurons,hawkins2019framework} & Discrete event-dr. & Sparse distrib.\ repr. & Sparse binary & Dendrite-local Hebbian & Arch.\ constr. & Local/alg. & No global gradient \\
Energy-based generative & RBM, DBN \cite{hinton2002training,hinton2006fast} & Discrete & Visible + hidden units & Dense real-valued & Contrastive divergence & None/Hist. & Local/alg. & Sampling cost \\
Static goodness-based & Forward-Forward \cite{hinton2022forward} & Static & None & Dense real-valued & Layerwise goodness objective & None/Hist. & Local/alg. & Layerwise objective generality \\
\bottomrule
\end{tabular}
\end{adjustbox}
\end{table}
 
\subsection{Comparison with Existing Surveys}
\label{sec:taxonomy_related}
 
Existing surveys (Table~\ref{tab:related_surveys}) cluster around three poles: the learning axis \cite{schmidgall2024brain,lillicrap2020backpropagation,ororbia2023brain,ororbia2024review,lv2024biologically,salvatori2025neuromimetic}, the spiking--neuromorphic column \cite{tavanaei2019deep,roy2019towards,schuman2022opportunities,rathi2023exploring,mazurek2025three,caviglia2026neurotrain,kudithipudi2025neuromorphic}, and continuous-time differential-equation models \cite{zhang2025surveyNODE}, with Hassabis et al.\ \cite{hassabis2017neuroscience} and Karniadakis et al.\ \cite{karniadakis2021physics} the broader exceptions. Each covers one or two combinations of Figure~\ref{fig:master_taxonomy} in depth and leaves the rest aside. The present survey is positioned across the joint state-dynamics and learning axes to make the disconnect visible across families that single-axis surveys treat separately.
 
\begin{table*}[!htbp]
\caption{Coverage of recent surveys against this paper's model-family categories. \(\bullet\): substantive; \(\circ\): partial; ---: out of scope. The distinguishing contribution of the current survey is not broader coverage alone but the specific joint decomposition it applies: state dynamics $\times$ credit assignment $\times$ biological grounding $\times$ substrate. This joint decomposition is what makes the forward--backward disconnect visible as a categorical pattern rather than as a scattered observation across independently organized reviews.}
\label{tab:related_surveys}
\centering
\tiny
\setlength{\tabcolsep}{3pt}
\renewcommand{\arraystretch}{0.85}
\begin{tabular}{@{}p{2.8cm} p{1.7cm} c c c c c c c@{}}
\toprule
\textbf{Survey} & \textbf{Venue, Year} & \makecell{\textbf{Static}} & \makecell{\textbf{Discrete}\\\textbf{/ seq.}} & \makecell{\textbf{Attn /}\\\textbf{SSM}} & \makecell{\textbf{Cont. /}\\\textbf{Implicit}} & \makecell{\textbf{Spiking}\\\textbf{/ hybrid}} & \makecell{\textbf{Bio}\\\textbf{learn.}} & \makecell{\textbf{Neuro-}\\\textbf{morph.}} \\
\midrule
Hassabis et al.\ \cite{hassabis2017neuroscience} & \emph{Neuron}, 2017 & \(\circ\) & \(\circ\) & --- & --- & \(\circ\) & \(\circ\) & --- \\
Tavanaei et al.\ \cite{tavanaei2019deep} & \emph{Neural Netw.}, 2019 & --- & --- & --- & --- & \(\bullet\) & \(\circ\) & \(\circ\) \\
Roy et al.\ \cite{roy2019towards} & \emph{Nature}, 2019 & --- & --- & --- & --- & \(\bullet\) & \(\circ\) & \(\bullet\) \\
Lillicrap et al.\ \cite{lillicrap2020backpropagation} & \emph{Nat.\ Rev.\ Neurosci.}, 2020 & --- & --- & --- & --- & --- & \(\bullet\) & --- \\
Karniadakis et al.\ \cite{karniadakis2021physics} & \emph{Nat.\ Rev.\ Phys.}, 2021 & \(\bullet\) & --- & --- & \(\circ\) & --- & --- & --- \\
Schuman et al.\ \cite{schuman2022opportunities} & \emph{Nat.\ Comput.\ Sci.}, 2022 & --- & --- & --- & --- & \(\bullet\) & \(\circ\) & \(\bullet\) \\
Rathi et al.\ \cite{rathi2023exploring} & \emph{ACM CSUR}, 2023 & --- & --- & --- & --- & \(\bullet\) & \(\circ\) & \(\bullet\) \\
Ororbia \cite{ororbia2023brain} & arXiv, 2023 & --- & \(\circ\) & --- & --- & \(\circ\) & \(\bullet\) & --- \\
Schmidgall et al.\ \cite{schmidgall2024brain} & \emph{APL Mach.\ Learn.}, 2024 & --- & \(\circ\) & --- & --- & \(\circ\) & \(\bullet\) & --- \\
Ororbia et al.\ \cite{ororbia2024review} & arXiv, 2024 & \(\circ\) & \(\circ\) & --- & \(\circ\) & \(\circ\) & \(\bullet\) & --- \\
Lv et al.\ \cite{lv2024biologically} & arXiv, 2024 & \(\circ\) & \(\circ\) & --- & --- & \(\circ\) & \(\bullet\) & \(\circ\) \\
Kudithipudi et al.\ \cite{kudithipudi2025neuromorphic} & \emph{Nature}, 2025 & --- & --- & --- & --- & \(\bullet\) & \(\circ\) & \(\bullet\) \\
Zhang et al.\ \cite{zhang2025surveyNODE} & PAKDD, 2025 & --- & --- & --- & \(\bullet\) & --- & --- & --- \\
Mazurek et al.\ \cite{mazurek2025three} & \emph{Patterns}, 2025 & --- & --- & --- & --- & \(\bullet\) & \(\bullet\) & \(\circ\) \\
Salvatori et al.\ \cite{salvatori2025neuromimetic} & \emph{Neural Netw.}, 2025/2026 & --- & --- & --- & --- & --- & \(\bullet\) & --- \\
Caviglia et al.\ \cite{caviglia2026neurotrain} & arXiv, 2026 & --- & --- & --- & --- & \(\bullet\) & \(\bullet\) & \(\circ\) \\
\textbf{Current survey} & This work & \(\bullet\) & \(\bullet\) & \(\bullet\) & \(\bullet\) & \(\bullet\) & \(\bullet\) & \(\bullet\) \\
\bottomrule
\end{tabular}
\end{table*}
 
\subsection{Literature Procedure and Scope Boundaries}
\label{sec:taxonomy_scope}
 
Literature was identified through ACM Digital Library, IEEE Xplore, Scopus, arXiv, and Google Scholar with citation chaining from foundational works, covering high-impact developments through 2026 (last comprehensive search June 2026). Supplementary Section~S1 documents the full procedure: databases, temporal windows, Boolean search strings grouped by taxonomy axis, inclusion/exclusion criteria, adjudication rule (Sections~\ref{sec:taxonomy_dynamics}--\ref{sec:taxonomy_bio} are authoritative), and procedural limitations. The atomic unit of classification is the architecture--learning configuration of Section~\ref{sec:taxonomy_learning}; placement rationales for ambiguous cases are documented in the supplementary material. Several active research programs sit outside scope: graph neural networks, Kolmogorov--Arnold networks, and GFlowNets are organized by structural inductive biases orthogonal to state-evolution form; diffusion models, variational autoencoders, and generative adversarial networks by generative objective rather than dynamics or neuroscience grounding; capsule networks, Neural Turing Machines, Differentiable Neural Computers, and world models by memory or environment-interaction objectives lying outside the state-dynamics taxonomy. None of these exclusions is a judgment about importance.
 
The configuration counts should be read carefully. They are descriptive of the audited representative set and not estimates of literature prevalence; coverage is conditioned on the search cutoff, indexing sources, scope rule, and classification criteria documented in Supplementary Section~S1. The taxonomy is intended as a reproducible analytical projection rather than an exhaustive census of neural-computation research, and the mechanism-level sensitivity analysis of Supplementary Section~S3 tests whether the qualitative asymmetry survives a finer decomposition of the backward axis.
 
\medskip
\noindent\textbf{Operational classification protocol.} To classify a new architecture--learning configuration under this taxonomy, we first identify the state variable and the mathematical rule by which it evolves, fixing the forward state-dynamics class of Section~\ref{sec:taxonomy_dynamics}. We then identify the mechanism that updates the parameters shaping those dynamics, fixing the primary credit-assignment class of Section~\ref{sec:taxonomy_learning}; distinct learning rules applied to the same forward operator are treated as distinct configurations, and the five secondary dimensions of Table~\ref{tab:secondary_dims} locate the mechanism within its primary class. Forward and learning biological grounding are then assigned independently using the tier definitions of Section~\ref{sec:taxonomy_bio}. Where more than one interpretation is defensible, the defining computational mechanism determines the primary placement and the alternative reading is recorded explicitly rather than creating an additional primary class. Readout-only training of fixed reservoirs is not counted as a configuration because the trained parameters do not shape the forward recurrent dynamics.

\section{Static Feedforward Input-Output Mappings}
\label{sec:static}
 
The Static class contains memoryless input-output maps of the form \(\mathbf{y} = f_\theta(\mathbf{x})\) from Eq.~\eqref{eq:intro_static_mapping}, in which no persistent internal state is carried across input examples. The class is dominated by Global gradient training, and the forward--backward disconnect is not yet diagnostic here because the forward dynamics are trivial; the class matters mainly as the baseline against which later sections measure departures along the state-dynamics axis.
 
\subsection{Perceptrons and Multilayer Networks}
\label{sec:static_mlp}
 
The perceptron \cite{rosenblatt1958perceptron} computes \(\mathbf{y} = \sigma(W\mathbf{x} + \mathbf{b})\). Its \gls{mlp} generalization \cite{rumelhart1986learning} composes the same affine-plus-nonlinearity map across layers, with the final-layer activation taken as the output, and is trained by backpropagation. In the state-evolution notation of Section~\ref{sec:foundations_state}, the layer index is analogous to a discrete time index, but no state is carried across input examples; this is what we mean by \emph{static}. The \gls{mlp} sits in the Global gradient row at biological grounding \emph{None or historical inspiration}: the perceptron's original motivation referenced information storage in the brain, but the resulting model retains essentially none of the temporal or biophysical structure of biological neurons identified in Section~\ref{sec:foundations}. The principal scalability bottleneck is the memory cost of storing per-layer activations for the backward pass.
 
\subsection{Convolutional Architectures: Neocognitron, CNN, ResNet}
\label{sec:static_cnn}
 
Convolutional architectures replace dense weight matrices with weight-shared local operators of the form \(\mathbf{h}^{(\ell+1)} = \sigma(W^{(\ell)} * \mathbf{h}^{(\ell)} + \mathbf{b}^{(\ell)})\), with \(*\) denoting convolution. Three families share this template but occupy different positions in Figure~\ref{fig:master_taxonomy}.
 
The Neocognitron \cite{fukushima1980neocognitron} predates the modern \gls{cnn} and derives its hierarchical simple/complex cell alternation from the receptive-field organization of cat primary visual cortex, trained by a local unsupervised competitive rule rather than by global backpropagation. It sits in the Local plasticity row at \emph{Architectural constraint} because its alternation directly mirrors a structural feature of visual cortex; depth is the price of this constraint, since the absence of a global gradient prevents scaling to backpropagation-trained \gls{cnn} depths. The modern \gls{cnn} \cite{lecun1998gradient} retains the convolutional structure but is trained by backpropagation, moving into the Global gradient row at \emph{None or historical inspiration}: the inductive bias was originally cortex-motivated, but the modern justification is statistical (translation equivariance, parameter sharing) rather than neurobiological.
 
\gls{resnet} \cite{he2016deep} adds an identity skip connection,
\begin{equation}
\mathbf{h}^{(\ell+1)} = \mathbf{h}^{(\ell)} + f\bigl(\mathbf{h}^{(\ell)};\, \theta^{(\ell)}\bigr),
\label{eq:static_resnet}
\end{equation}
which formally identifies the layer update with a discrete Euler step of \(\dot{\mathbf{h}}(t) = f(\mathbf{h}(t); \theta(t))\). The static taxonomy combination containing \gls{resnet} is therefore structurally adjacent to the continuous-time column, the basis for the Neural ODE reading developed in Section~\ref{sec:continuous}, although \gls{resnet} itself remains a finite-layer static map at training and inference time. Biological grounding is \emph{None or historical inspiration}.
 
\subsection{Physics-Informed Neural Networks}
\label{sec:static_pinn}
 
\glspl{pinn} \cite{raissi2019physics,karniadakis2021physics} are static feedforward networks trained to approximate solutions of ordinary or partial differential equations. The network \(u_\theta(\mathbf{x}, t)\) is treated as an ansatz for the solution \(u(\mathbf{x}, t)\), and the training objective combines a data term with a residual term that penalizes violation of the governing equation:
\begin{equation}
\mathcal{L}(\theta) = \mathcal{L}_{\mathrm{data}}(\theta) + \lambda\, \mathcal{L}_{\mathrm{residual}}(\theta),
\qquad
\mathcal{L}_{\mathrm{residual}}(\theta) = \frac{1}{N_r} \sum_{i=1}^{N_r} \bigl\lVert \mathcal{D}\, u_\theta(\mathbf{x}_i, t_i) \bigr\rVert^2,
\label{eq:static_pinn_loss}
\end{equation}
with \(\mathcal{D}\) the differential operator defining the \gls{pde}, the residual evaluated at collocation points \((\mathbf{x}_i, t_i)\), and \(\lambda\) balancing the two terms. Derivatives required by \(\mathcal{D}\) are computed by automatic differentiation through the network itself.
 
The taxonomic placement of \glspl{pinn} is informative. The task is dynamical because \(u(\mathbf{x}, t)\) solves a differential equation that describes physical evolution, but the model is a static feedforward map from coordinates to field values, with no persistent internal state and no recurrence. Dynamics enter only through the loss. A static model can therefore solve a dynamical problem, provided the dynamics are encoded into the training objective rather than into the state evolution. The cost is paid elsewhere: \gls{pinn} training requires careful balancing of \(\lambda\), is sensitive to the conditioning of the residual loss, and tends to struggle with stiff or multi-scale problems \cite{karniadakis2021physics}. Biological grounding is \emph{None or historical inspiration}.
 
\subsection{Section Synthesis}
\label{sec:static_reflection}
 
The Static class is dominated by Global gradient training, with the Neocognitron (Local plasticity, \emph{Architectural constraint}) and Forward-Forward (Energy/unsupervised) as exceptions. Two structural moves of later sections originate here: the \gls{resnet} skip connection (Eq.~\eqref{eq:static_resnet}) is the discrete Euler step that Neural ODEs lift to a continuous-time vector field (Section~\ref{sec:continuous}), and the \gls{pinn} residual loss (Eq.~\eqref{eq:static_pinn_loss}) is the engineering precedent for encoding dynamical constraints into a non-dynamical model, an idea that returns with energy-based objectives in Section~\ref{sec:learning}. Biologically the class is shallow: convolutional weight sharing is a translation-invariance prior rather than a derivation from cortical column organization, with the Neocognitron the exception that makes the simple/complex cell borrowing literal.

\section{Discrete-Time Recurrent Dynamics}
\label{sec:discrete}
 
Model families with an explicit hidden state \(\mathbf{h}_t\) updated by a difference equation occupy the Discrete-time class, recovering Eq.~\eqref{eq:state_evolution_discrete}. The introduction of state changes what learning has to do: a static network's backward pass credit-assigns across layers, but a recurrent network's backward pass must credit-assign across time. This is where the forward--backward disconnect first becomes visible. Transformer and structured \glspl{ssm} share this class. Modern Hopfield is treated alongside them in Section~\ref{sec:attention_ssm} because it provides the associative-memory interpretation of attention, despite belonging to the Implicit class.
 
\subsection{Classical Recurrent Neural Networks}
\label{sec:discrete_rnn}
 
The simple \gls{rnn} \cite{elman1990finding} computes
\begin{equation}
\mathbf{h}_t = \sigma\bigl(W_h \mathbf{h}_{t-1} + W_x \mathbf{x}_t + \mathbf{b}\bigr),
\qquad
\mathbf{y}_t = W_y \mathbf{h}_t + \mathbf{b}_y,
\label{eq:discrete_rnn}
\end{equation}
where \(\mathbf{h}_t \in \mathbb{R}^d\) is the hidden state at step \(t\), \(W_h, W_x, W_y\) are learnable weight matrices, and \(\sigma\) is an element-wise nonlinearity. Eq.~\eqref{eq:discrete_rnn} is the discrete-time analogue of the continuous-time state-evolution equation Eq.~\eqref{eq:state_evolution_continuous}, with the same parameter set \(\theta_F = \{W_h, W_x, \mathbf{b}\}\) reused at every time step. Parameter sharing across time is a real but loose biological analogy: a single recurrent operator applied repeatedly mirrors the temporal homogeneity of cortical circuits at a functional level. Under the split-axis grounding scheme of Section~\ref{sec:taxonomy_bio}, the \gls{rnn}'s forward grounding is \emph{Functional analogy} rather than \emph{Architectural constraint}: the recurrent operator would have been adopted independently for sequence modeling, so biology does not rule out alternative design choices the way the Neocognitron's specific simple/complex alternation does. The learning grounding is \emph{None} because the rule is \gls{bptt}. The units themselves remain point neurons rather than biophysical models, omitting membrane dynamics, dendritic integration, multiple time constants, and neuromodulation.
 
\glspl{rnn} are trained by \gls{bptt} \cite{werbos1990backpropagation}, which unrolls the recurrence into a layered graph with tied weights and applies backpropagation. The resulting gradient is exact but its magnitude depends on a product of Jacobians of \(\sigma\) and \(W_h\) across time steps, and tends to vanish or explode as sequence length grows, with the largest singular value of \(W_h\) controlling which regime is reached \cite{bengio1994learning,pascanu2013difficulty}. The vanishing-gradient regime makes long-range temporal credit assignment effectively impossible without architectural intervention; the exploding regime can be mitigated by gradient clipping. The \gls{rnn} sits cleanly in the Global gradient category but reveals the first tension the survey develops: a learning rule exact in principle becomes hard to apply once the forward computation is dynamical.
 
\subsection{Gated Recurrence: LSTM and GRU}
\label{sec:discrete_gated}
 
Gated recurrent architectures address the vanishing-gradient problem by introducing additional state variables and multiplicative gates that regulate information flow. The \gls{lstm} \cite{hochreiter1997long} maintains a cell state \(\mathbf{c}_t\) alongside the hidden state \(\mathbf{h}_t\), and computes three sigmoid gates \(\mathbf{f}_t, \mathbf{i}_t, \mathbf{o}_t = \sigma_g(W_{f,i,o}\,[\mathbf{h}_{t-1}, \mathbf{x}_t] + \mathbf{b}_{f,i,o})\) from the concatenation \([\mathbf{h}_{t-1}, \mathbf{x}_t]\) to control forgetting, input integration, and output. The analytically relevant piece is the cell-state additive update,
\begin{equation}
\mathbf{c}_t = \mathbf{f}_t \odot \mathbf{c}_{t-1} + \mathbf{i}_t \odot \tilde{\mathbf{c}}_t,
\label{eq:discrete_lstm}
\end{equation}
where \(\tilde{\mathbf{c}}_t = \tanh(W_c [\mathbf{h}_{t-1}, \mathbf{x}_t] + \mathbf{b}_c)\) is the candidate update, \(\odot\) is element-wise multiplication, and the hidden state is read out as \(\mathbf{h}_t = \mathbf{o}_t \odot \tanh(\mathbf{c}_t)\). The \gls{gru} \cite{cho2014learning} collapses the cell state into the hidden state and uses two gates (reset and update), giving a more compact recurrence with comparable empirical performance.
 
The dynamical-systems interpretation of the gates is the relevant content for the taxonomy. Without gating, Eq.~\eqref{eq:discrete_rnn} has a contracting or expanding Jacobian determined by \(\sigma'(\cdot) W_h\), producing the vanishing or exploding gradient regimes \cite{pascanu2013difficulty}. The \gls{lstm} cell-state update admits a Jacobian whose diagonal is dominated by the forget-gate values \(\mathbf{f}_t\), so when these values are near one the gradient passes through essentially unattenuated. Gating is thus best read as a learned modulation of the recurrent Jacobian that keeps the trained dynamics close to a stable but information-preserving regime, anticipating the Jacobian-stability arguments of Section~\ref{sec:continuous}.
 
Biologically, gating is functionally analogous to multiplicative modulation by neuromodulators or shunting inhibition, though the mapping is qualitative rather than constraint-derived: the artificial gate is a learned scalar applied uniformly across a hidden dimension and omits spatial neurotransmitter release, receptor chemistry, and time-varying neuromodulator availability. \glspl{lstm} and \glspl{gru} accordingly sit at \emph{Functional analogy} on the forward axis and \emph{None} on the learning axis.
 
\subsection{Classical Hopfield Networks}
\label{sec:discrete_hopfield}
 
The classical Hopfield network \cite{hopfield1982neural} is a fully connected recurrent network of binary or bipolar units evolved by the asynchronous update \(h_{i,t+1} = \mathrm{sgn}(\sum_j W_{ij}\, h_{j,t} - \theta_i)\) with a symmetric weight matrix \(W_{ij} = W_{ji}\) and zero self-couplings. The defining property is that this dynamics is a gradient descent on the Lyapunov energy
\begin{equation}
E(\mathbf{h}) = -\tfrac{1}{2} \sum_{i,j} W_{ij}\, h_i\, h_j + \sum_i \theta_i\, h_i,
\label{eq:discrete_hopfield_energy}
\end{equation}
guaranteeing convergence to one of a finite set of fixed-point attractors. Memories are encoded as the stable patterns of \(E\), and recall is implemented by initializing the network with a corrupted version of a stored pattern and letting the dynamics relax to the nearest attractor.
 
Learning is Hebbian: the storage rule is \(W_{ij} \propto \sum_\mu \xi_i^\mu \xi_j^\mu\) for a set of patterns \(\{\xi^\mu\}\), a single-shot local plasticity update rather than \gls{bptt}. This places classical Hopfield in the Local plasticity category at biological grounding \emph{Architectural constraint}: attractor dynamics over a symmetric weight matrix is a structural commitment derived from associative-memory biology that the model would not otherwise adopt. The principal limitation is capacity (only \(O(N)\) patterns for \(N\) units before retrieval breaks down, with a spurious-state landscape), substantially relaxed by the modern Hopfield reformulation \cite{ramsauer2021hopfield} treated with attention and structured \glspl{ssm} in Section~\ref{sec:attention_ssm}.
 
\subsection{Trained RNNs as Dynamical Systems}
\label{sec:discrete_trained_dynamics}
 
\begin{figure}[t]
\centering
\begin{tikzpicture}[
  axis style/.style = {->, >=stealth, line width=0.5pt, black!50},
  flow/.style       = {->, >=stealth, line width=0.6pt, black!75},
  fp dot/.style     = {circle, fill=black, inner sep=1.6pt},
  separatrix/.style = {dashed, line width=0.4pt, black!40},
  panel label/.style = {font=\footnotesize\bfseries, anchor=south},
  panel caption/.style = {font=\footnotesize\itshape, anchor=north, align=center},
  axis label/.style  = {font=\scriptsize, anchor=north east}
]
 
\begin{scope}[shift={(0,0)}]
  \draw[axis style] (-1.4,-1.4) -- (1.4,-1.4) node[axis label] {\(h_1\)};
  \draw[axis style] (-1.4,-1.4) -- (-1.4,1.4) node[above, font=\scriptsize] {\(h_2\)};
  \node[fp dot] (pA) at (0,0) {};
  \foreach \ang in {0,45,90,135,180,225,270,315} {
    \draw[flow] ({1.05*cos(\ang)},{1.05*sin(\ang)}) -- ({0.32*cos(\ang)},{0.32*sin(\ang)});
  }
  \node[panel label] at (0, 1.55) {(a) Point attractor};
  \node[panel caption] at (0, -1.65) {stable fixed point\\(memory storage)};
\end{scope}
 
\begin{scope}[shift={(4.6,0)}]
  \draw[axis style] (-1.4,-1.4) -- (1.4,-1.4) node[axis label] {\(h_1\)};
  \draw[axis style] (-1.4,-1.4) -- (-1.4,1.4) node[above, font=\scriptsize] {\(h_2\)};
  \draw[line width=1.2pt, black] (-1.0,0) -- (1.0,0);
  \node[fp dot, scale=0.7] at (-0.7,0) {};
  \node[fp dot, scale=0.7] at (0,0) {};
  \node[fp dot, scale=0.7] at (0.7,0) {};
  \foreach \x in {-0.9,-0.45,0,0.45,0.9} {
    \draw[flow] (\x, 0.95) -- (\x, 0.18);
    \draw[flow] (\x,-0.95) -- (\x,-0.18);
  }
  \node[panel label] at (0, 1.55) {(b) Line attractor};
  \node[panel caption] at (0, -1.65) {one-parameter family of stable states\\(graded integration)};
\end{scope}
 
\begin{scope}[shift={(9.2,0)}]
  \draw[axis style] (-1.4,-1.4) -- (1.4,-1.4) node[axis label] {\(h_1\)};
  \draw[axis style] (-1.4,-1.4) -- (-1.4,1.4) node[above, font=\scriptsize] {\(h_2\)};
  \draw[separatrix] (-1.05,-1.05) -- (1.05,1.05);
  \draw[separatrix] (-1.05, 1.05) -- (1.05,-1.05);
  \node[fp dot] (pC) at (0,0) {};
  \draw[flow] (-1.0,-1.0) -- (-0.32,-0.32);
  \draw[flow] ( 1.0, 1.0) -- ( 0.32, 0.32);
  \draw[flow] (-0.32, 0.32) -- (-1.0, 1.0);
  \draw[flow] ( 0.32,-0.32) -- ( 1.0,-1.0);
  \node[panel label] at (0, 1.55) {(c) Saddle point};
  \node[panel caption] at (0, -1.65) {unstable fixed point\\(decision boundary)};
\end{scope}
 
\end{tikzpicture}
\caption{Three canonical dynamical objects recovered in trained \gls{rnn} state spaces by the fixed-point reverse-engineering technique of \cite{sussillo2013opening}. Solid black dots mark fixed points; arrows indicate the local flow of the dynamics in a 2D projection \((h_1, h_2)\) of the hidden state. A \emph{point attractor} (a) implements memory storage. A \emph{line attractor} (b) implements graded integration along a one-parameter family of stable states and is the canonical mechanism for context-dependent decision making in trained \glspl{rnn} and in prefrontal cortex \cite{mante2013context}. A \emph{saddle point} (c) separates basins of attraction (dashed separatrices) and implements a decision boundary between two stored states. Across populations of trained networks the precise weights differ but the topology of these objects is conserved \cite{maheswaranathan2019universality}.}
\Description{Three side-by-side phase-portrait panels in a 2D projection of recurrent-network hidden state. Panel (a) shows a point attractor: a single filled dot in the center with arrows pointing inward from all directions. Panel (b) shows a line attractor: a horizontal line of stable states with arrows pointing toward the line from above and below. Panel (c) shows a saddle point: a central dot with arrows along one axis flowing inward and along the perpendicular axis flowing outward, with dashed separatrices dividing the plane into basins of attraction.}
\label{fig:attractors}
\end{figure}
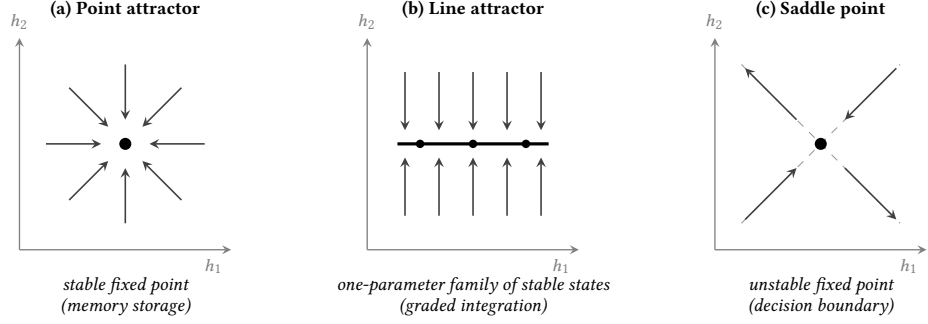

A line of work beginning with Sussillo and Barak \cite{sussillo2013opening} treats a trained recurrence as a dynamical system rather than as a black-box function approximator: candidate fixed points are located by minimizing \(q(\mathbf{h}) = \tfrac{1}{2}\lVert F(\mathbf{h}, \mathbf{x}; \theta_F) - \mathbf{h} \rVert^2\), and the Jacobian \(\partial F / \partial \mathbf{h}\) at each candidate determines local stability and direction structure. Applied to bit-flip-flop storage, sine-wave generation, and context-dependent decision making \cite{sussillo2013opening,mante2013context}, the procedure recovers the small repertoire of low-dimensional dynamical objects shown in Figure~\ref{fig:attractors}.
 
Across populations of trained networks the precise recurrent weights differ widely but the topology of the fixed-point structure is conserved \cite{maheswaranathan2019universality}: networks trained on the same task converge to dynamically equivalent solutions. This universality empirically supports treating \glspl{rnn} as dynamical systems rather than opaque function approximators, which is the premise of the taxonomy in Section~\ref{sec:taxonomy}, and foreshadows the Jacobian-based stability analyses of the continuous-time models in Section~\ref{sec:continuous}. Trained \glspl{rnn} have also been used as models of prefrontal cortex performing context-dependent integration \cite{mante2013context}, which is why the \gls{rnn} entry sits at \emph{Functional analogy} in Figure~\ref{fig:master_taxonomy}: the same recurrent operator is applied at every time step in a way that matches the temporal homogeneity assumed in computational neuroscience models, while omitting slow currents, dendritic compartments, and short-term synaptic dynamics.
 
\subsection{Section Synthesis}
\label{sec:discrete_reflection}
 
The Discrete-time class spans all four credit-assignment categories: Global gradient (\gls{rnn}, \gls{lstm}, \gls{gru} trained by \gls{bptt}); Approximate or implicit gradient (\gls{rnn} with synthetic gradients \cite{jaderberg2017decoupled} and with RFLO random-feedback local online learning \cite{murray2019local}); Local plasticity (classical Hopfield and hierarchical temporal memory with dendrite-local Hebbian rules); and Energy/unsupervised (restricted Boltzmann machines and deep belief networks trained by contrastive divergence). This is the first class in the traversal where the forward--backward disconnect becomes diagnostic: the same forward architecture supports radically different credit-assignment mechanisms, and the choice of mechanism rather than the forward dynamics determines whether the model scales.
 
The trained-\gls{rnn} dynamical-systems literature shows that successfully trained recurrent networks converge to a small repertoire of low-dimensional dynamical objects (point attractors, line attractors, slow manifolds) analyzable quantitatively; the same vocabulary that describes a \gls{lif} neuron and a Neural ODE also describes what trained \glspl{rnn} actually do. Across the column, stronger biological preservation correlates with weaker conventional engineering scalability: the \gls{rnn} preserves only recurrent-loop topology (omitting membrane dynamics, dendrites, and neuromodulation) and is compensated by \gls{bptt}; \glspl{lstm} and \glspl{gru} add the functional analogue of multiplicative neuromodulation via learned scalar gates; the classical Hopfield network preserves attractor dynamics and Hebbian storage but pays a capacity ceiling at \(O(N)\) patterns.

\section{Attention, Modern Hopfield, and Structured State-Space Models}
\label{sec:attention_ssm}
 
Attention, modern Hopfield retrieval, and structured \glspl{ssm} are treated together in this section because they provide three related mechanisms for sequence memory: direct content-based retrieval, associative-memory retrieval, and compressed recurrent state evolution. They are not the same architecture, but recent equivalence results expose formal bridges among them. Modern Hopfield retrieval recovers softmax self-attention under a specific single-step update when the inverse-temperature parameter is set to \(\beta = 1/\sqrt{d_k}\) \cite{ramsauer2021hopfield}, and structured state-space duality connects a class of \glspl{ssm} with structured-masked attention and linear-attention-like recurrences \cite{dao2024ssd,katharopoulos2020linear}. Modern Hopfield retrieval belongs to the Implicit class; its associative-memory interpretation is the analytical bridge that connects attention to the engineered sequence models of this class, which motivates its inclusion here.
 
\subsection{Self-Attention as Content-Dependent Retrieval}
\label{sec:attention_ssm_attention}
 
Attention implements a soft read from a memory whose contents are themselves a function of the input sequence. Introduced by Bahdanau et al.\ \cite{bahdanau2015neural} as an additive soft alignment for translation and reformulated as scaled dot product by Vaswani et al.\ \cite{vaswani2017attention},
\begin{equation}
\mathrm{Attention}(Q, K, V) = \mathrm{softmax}\!\left(\frac{Q K^{\top}}{\sqrt{d_k}}\right) V,
\label{eq:attention}
\end{equation}
Eq.~\eqref{eq:attention} is the canonical differentiable key-value retrieval object: each query draws a convex combination of value rows weighted by query-key similarity. During parallel training, no persistent dynamical state is carried between positions; autoregressive inference introduces a key-value cache that functions as an expanding external state. Forward grounding is \emph{None or historical inspiration} and learning grounding is \emph{None}. The conventional scalability bottleneck is quadratic dependence of \(Q K^{\top}\) on sequence length, although activation memory can be reduced to linear via tiled exact-attention implementations such as FlashAttention-3 \cite{dao2024flash3}.
 
\subsection{Modern Hopfield Retrieval}
\label{sec:attention_ssm_hopfield}
 
The associative-memory reading of attention is made precise by the continuous modern Hopfield network of Ramsauer et al.\ \cite{ramsauer2021hopfield}. Earlier dense associative memories \cite{krotov2016dense,demircigil2017model} replaced the quadratic energy of the classical Hopfield network \cite{hopfield1982neural} with polynomial and exponential interaction functions, raising the storage capacity from \(O(N)\) patterns to \(O(d^{n-1})\) and \(2^{d/2}\) respectively for \(N\) patterns of dimension \(d\). The continuous version of \cite{ramsauer2021hopfield} adopts the log-sum-exp energy
\begin{equation}
E(\boldsymbol{\xi}) = -\,\mathrm{lse}(\beta, X^{\top}\boldsymbol{\xi}) + \tfrac{1}{2}\boldsymbol{\xi}^{\top}\boldsymbol{\xi} + \beta^{-1}\log N + \tfrac{1}{2}M^{2},
\label{eq:hopfield_energy}
\end{equation}
where \(X = (\mathbf{x}_1, \ldots, \mathbf{x}_N)\) collects stored patterns, \(\boldsymbol{\xi}\) is the network state, and \(M = \max_i \lVert \mathbf{x}_i \rVert\). The Concave-Convex Procedure on \(E\) yields the single-step update \(\boldsymbol{\xi}_{\mathrm{new}} = X \cdot \mathrm{softmax}(\beta X^{\top} \boldsymbol{\xi})\), and the resulting storage capacity is exponential in the pattern dimension under the separation assumptions of Ramsauer et al.\ \cite{ramsauer2021hopfield}.
 
Identifying \(X = K\), \(\boldsymbol{\xi} = Q\), \(V = X W_V\) and projecting through value weights gives the attention-equivalence theorem of Ramsauer et al.\ (their Eq.~10): at \(\beta = 1/\sqrt{d_k}\), one Hopfield update step is exactly one attention layer in the form of Eq.~\eqref{eq:attention}, and the same step achieves single-step retrieval with error that is exponentially small in the pattern-separation margin. The forward grounding is \emph{Functional analogy} because the attractor-and-energy interpretation is the structural analogue to cortical associative memory at a functional level rather than a constraint-derived adoption of a specific biological mechanism, and the learning grounding is \emph{None} when modern Hopfield is embedded as a layer in an end-to-end backpropagation-trained network. The Energy Transformer \cite{hoover2023energy} makes the unification constructive by reformulating the transformer block as gradient descent on a single engineered Hopfield-style energy.
 
\subsection{Structured State-Space Models}
\label{sec:attention_ssm_ssm}
 
Structured \glspl{ssm} treat the sequence as the output of a linear (or selectively-parameterized) state-space system. The continuous-time form \(\mathbf{h}'(t) = A \mathbf{h}(t) + B \mathbf{x}(t),\ \mathbf{y}(t) = C \mathbf{h}(t)\) is discretized through a bilinear or zero-order-hold rule, giving a recurrence \(\mathbf{h}_t = \bar{A} \mathbf{h}_{t-1} + \bar{B} \mathbf{x}_t\). The HiPPO framework \cite{gu2020hippo} chooses \(A\) so that the hidden state maintains an online compression of the input history onto an orthogonal-polynomial basis; this initialization underlies S4 \cite{gu2022efficiently}, which combines a diagonal-plus-low-rank decomposition of \(A\) with a parallelizable convolutional view and achieves long-context performance on tasks where Transformer-class models fail (Path-X at length 16{,}384). S5 \cite{smith2023s5} simplifies the construction to a diagonalized multi-input multi-output \gls{ssm} with a parallel scan. Mamba \cite{gu2023mamba} breaks linear time-invariance by making \(B\), \(C\), and the discretization step \(\Delta\) input-dependent, matching Transformer-class language-modeling performance up to the 3B-parameter scale, although pure-\gls{ssm} variants lag the Transformer on copying and in-context retrieval \cite{waleffe2024empirical}.
 
State-space duality \cite{dao2024ssd} establishes that structured \glspl{ssm} are equivalent to semiseparable matrices, that linear attention generalizes to structured-masked attention, and that the intersection admits both an \gls{ssm}-like recurrent form and an attention-like quadratic form, reducing Mamba-2 at scalar-times-identity state transition to causal linear attention \cite{katharopoulos2020linear}. RWKV \cite{peng2023rwkv} and xLSTM \cite{beck2024xlstm} occupy the same neighborhood, the former with a linear time-mixing operator (0.1B--14B parameters), the latter with matrix-memory and scalar-memory variants whose update mirrors fast-weight memory. The forward grounding for all structured \glspl{ssm} is \emph{None or historical inspiration} (control-theoretic system identification, not biological computation); the learning grounding is \emph{None}. The scalability bottleneck is state-update parallelism on target hardware, and pure-\gls{ssm} architectures so far underperform on tasks requiring explicit long-context retrieval \cite{waleffe2024empirical}.
 
\subsection{Section Synthesis}
\label{sec:attention_ssm_synthesis}
 
The three families do not collapse into a single architecture, but they occupy a common analytical neighborhood: all three address long-range sequence dependence by controlling how past information is stored, retrieved, or compressed. Modern Hopfield gives an associative-memory interpretation of softmax attention; linear attention shows that certain attention mechanisms admit an equivalent recurrent form \cite{katharopoulos2020linear}; and structured state-space duality shows that particular \glspl{ssm} and structured-masked attention share a semiseparable representation \cite{dao2024ssd}. The common thread is therefore not identity but a set of formal bridges between retrieval, recurrence, and compressed state, each holding only under the specific regime conditions stated by its source result.
 
In biological terms, Modern Hopfield preserves more of the cortical solution to content-addressable recall than the other two families by retaining a continuous attractor with an energy landscape under which retrieval is a single gradient step. Attention preserves the soft, content-addressable interface but discards the energy landscape, replacing it with a learnable feed-forward read. Structured \glspl{ssm} preserve neither and compress the sequence into a recurrent state through engineering-derived transition operators. The omissions across all three families are compensated by gradient-trainability and parallelism, which enable scaling beyond what biologically grounded alternatives have so far reached.
 
\section{Continuous-Time and Implicit Neural Dynamics}
\label{sec:continuous}
 
Continuous-time and implicit models define forward computation as the solution of a dynamical or equilibrium problem rather than the output of a finite stack of layers: a continuous-time trajectory in CTRNNs, Neural ODEs, and Liquid Neural Networks, or a fixed point in Deep Equilibrium Models. The shared issue is not primarily biological plausibility but state representation, and the analytical interest of the column is the way credit assignment is recovered through solver-defined dynamics rather than through a stored layerwise computation graph. The internal label \emph{solver-defined neural computation} captures this shift.
 
\subsection{Continuous-Time State Evolution: CTRNNs, Neural ODEs, and Liquid Networks}
\label{sec:continuous_state}
 
Continuous-time recurrent networks evolve a hidden state according to the leaky-integrator equation
\begin{equation}
\tau_i\, \frac{d h_i}{d t} = -h_i + \sum_{j} W_{ij}\, \sigma(h_j) + x_i(t),
\label{eq:ctrnn}
\end{equation}
with time constants \(\tau_i\), weights \(W_{ij}\), pointwise nonlinearity \(\sigma\), and input \(x_i(t)\). Funahashi and Nakamura \cite{funahashi1993approximation} proved that such networks can approximate any finite-time trajectory of an arbitrary \(n\)-dimensional dynamical system given enough hidden units and appropriate parameters. Chen et al.\ \cite{chen2018neural} reformulated this as the Neural ODE,
\begin{equation}
\frac{d \mathbf{h}(t)}{d t} = f(\mathbf{h}(t), t, \theta),
\label{eq:neural_ode}
\end{equation}
with \(f\) a learnable network and \(\mathbf{h}(t_1)\) computed from \(\mathbf{h}(t_0)\) by a numerical solver. The construction is the continuous-depth limit of a \gls{resnet}: the residual update of Eq.~\eqref{eq:static_resnet} becomes Eq.~\eqref{eq:neural_ode} as the step size shrinks to zero. Well-posedness is usually guaranteed when \(f\) is Lipschitz continuous, though numerical behavior still depends on solver tolerance and stiffness.
 
Liquid Time-Constant networks \cite{hasani2021liquid} couple the effective time constant to state and input,
\begin{equation}
\frac{d \mathbf{h}(t)}{d t} = -\left[\frac{1}{\tau} + f(\mathbf{h}, \mathbf{x}, t, \theta)\right] \mathbf{h}(t) + f(\mathbf{h}, \mathbf{x}, t, \theta) \odot A,
\label{eq:ltc}
\end{equation}
with bounded \(f\) and constant \(A\); state and time constant remain bounded on any finite interval, and a universality result establishes greater expressivity than CTRNNs and Neural ODEs sharing the same \(f\). The closed-form variant \cite{hasani2022closed} replaces the numerical solver with an analytical approximation for at least an order of magnitude speedup on continuous-control benchmarks; a 19-neuron variant derived from the \emph{C.\ elegans} connectome \cite{lechner2020neural} learns end-to-end visual lane-keeping, though the biological motivation is functional-analogy at the single-neuron level. Forward grounding is \emph{Functional analogy} for CTRNNs and Liquid networks and \emph{None or historical inspiration} for Neural ODEs; learning grounding is uniformly \emph{None}.
 
\subsection{Non-Unrolled Credit Assignment: Adjoint and Implicit Differentiation}
\label{sec:continuous_credit}
 
Continuous and implicit models do not eliminate global credit assignment: they replace stored backpropagation through a finite graph with gradients recovered through solver-defined dynamics. For the Neural ODE, Chen et al.\ \cite{chen2018neural} introduce an adjoint backward ODE for the state-gradient \(\mathbf{a}(t) = \partial L / \partial \mathbf{h}(t)\), giving memory cost constant in solver steps. The property is regime-dependent: Gholami et al.\ \cite{gholami2019anode} showed reverse-time reconstruction can be numerically inconsistent with the forward solve under common activations, and restoring accuracy by checkpointing requires \(O(L)\) memory.
 
Deep Equilibrium Models \cite{bai2019deep} take the construction to the fixed-point regime, with forward computation defined as the solution of
\begin{equation}
\mathbf{z}^{\star} = f_\theta(\mathbf{z}^{\star}, \mathbf{x}),
\label{eq:deq_fixedpoint}
\end{equation}
found by root-finding (Broyden or Anderson acceleration). The backward pass uses implicit differentiation: the implicit function theorem at the fixed point yields the parameter gradient as a single linear solve through \((I - \partial f / \partial \mathbf{z}^{\star})\), requiring only vector-Jacobian products, and activation memory is independent of effective depth (on WikiText-103, 172M-parameter DEQ-Transformer matches Transformer-XL perplexity at roughly one-third the activation memory \cite{bai2019deep}). Equivalence to a weight-tied infinite-depth network holds \emph{under the convergence assumption} that Eq.~\eqref{eq:deq_fixedpoint} admits a unique fixed point.
 
The Neural ODE adjoint and DEQ implicit differentiation are distinct machinery, but scalability for both depends on numerical stability rather than expressivity alone: adjoint accuracy degrades under stiffness, and the conditioning of the equilibrium Jacobian governs the noise floor of DEQ gradients; explicit Jacobian regularization \cite{bai2021stabilizing} recovers a constant-memory DEQ at approximately \gls{resnet}-101 speed and accuracy.
 
\subsection{Section Synthesis}
\label{sec:continuous_synthesis}
 
Continuous-time and implicit models change the state representation of neural computation but not its dominant credit-assignment mechanism. Forward computation moves from a finite stack of layers to the solution of a dynamical or equilibrium problem, gradients are recovered through solver-defined dynamics, and the conditioning of those backward equations becomes the operative scalability constraint in place of the memory and depth constraints of the Discrete-time class. Biologically, Neural ODEs, CTRNNs, and Liquid networks preserve continuous state evolution at the single-unit level; Deep Equilibrium Models preserve attractor-style fixed-point computation at the population level. Continuous-attractor models of grid cells \cite{burak2009accurate} furnish a parallel rather than an equivalence: stable biological state maintenance also depends on delicate dynamical conditions, similar in spirit to the Jacobian-conditioning constraint of implicit-differentiation backward passes. Forward grounding spans \emph{Functional analogy} and \emph{None or historical inspiration}; learning grounding is uniformly \emph{None}.
 
\section{Hybrid Event-Driven Systems}
\label{sec:spiking}
 
Hybrid event-driven systems constitute the Hybrid class and provide the strongest forward biological grounding in the survey: spiking neural networks restore membrane-potential dynamics, threshold-triggered discrete spikes, sparse binary communication, and temporal coding, placing the forward axis at \emph{Mechanistic plausibility}. The learning axis, however, splits sharply between surrogate-gradient training (which drives most current scaling but relies on a smoothed backward approximation) and biologically observed local plasticity (mechanistic learning grounding at smaller demonstrated scale). This class does not resolve the forward--backward disconnect; it makes the disconnect most directly visible.
 
\subsection{Spiking Forward Dynamics}
\label{sec:spiking_forward}
 
The third-generation framework of Maass \cite{maass1997networks} organizes neural-network computation into spiking networks that compute through discrete spike timing, distinguished from first-generation binary-output perceptrons and second-generation continuous-output gradient-trained units; spiking networks under temporal coding can simulate arbitrary feedforward sigmoidal networks of comparable size.
 
The forward computation of a spiking network is dominated by integrate-fire-reset dynamics at the single-neuron level: a \gls{lif} neuron maintains a continuous subthreshold membrane potential and emits a discrete spike \(S = \Theta(V - \vartheta)\) at threshold crossing, followed by reset. These three primitives (continuous subthreshold integration, threshold-triggered binary event emission, instantaneous reset) restore four properties absent from the model families of Sections~\ref{sec:static}--\ref{sec:continuous}: continuous-time membrane state, all-or-none communication, sparse binary coding, and temporal coding in which spike timing carries information beyond firing rate \cite{yamazaki2022spiking,tavanaei2019deep}.
 
The placement at \emph{Mechanistic plausibility} on the forward axis applies at the single-neuron level. The \gls{lif} model and threshold dynamics are biophysically motivated reductions of the Hodgkin-Huxley equations of Section~\ref{sec:foundations_neurons}, but population-level phenomena such as excitation-inhibition balance, dendritic integration, and neuromodulation remain idealized in standard \gls{snn} architectures. Energy and communication advantages of event-driven computation are deferred to the substrate discussion of Section~\ref{sec:neuromorphic}.
 
\subsection{Training Spiking Networks: Surrogate Gradients and Local Plasticity}
\label{sec:spiking_credit}
 
The same spiking forward architecture admits two qualitatively different training regimes that split sharply on the learning-grounding axis.
 
The dominant approach for scaling \glspl{snn} to large tasks is the surrogate-gradient method of Neftci, Mostafa, and Zenke \cite{neftci2019surrogate}. The forward pass preserves the non-differentiable spike \(S = \Theta(V - \vartheta)\), whose derivative is zero almost everywhere and ill-defined at threshold and therefore blocks standard gradient propagation. The backward pass substitutes the spike's derivative with a smooth surrogate,
\begin{equation}
\frac{\partial S}{\partial V}\ \longrightarrow\ \sigma'(V - \vartheta),
\label{eq:surrogate}
\end{equation}
where \(\sigma'\) is a sigmoid, fast-sigmoid, triangular, or super-spike-style derivative. Training is gradient descent on what Neftci et al.\ describe as a virtual surrogate loss \cite{neftci2019surrogate}: the backward pass computes the gradient of the smoothed surrogate training system, not the true derivative of the original spike discontinuity, and the forward computation remains spiking. Specific choices of \(\sigma'\) exist but do not change the taxonomic placement, since all remain global gradient methods applied through a smoothed backward approximation.
 
Surrogate-gradient training has driven the scaling of directly-trained spiking networks to ImageNet-class benchmarks. The spike-element-wise residual construction of Fang et al.\ \cite{fang2021deep} enabled directly-trained \glspl{snn} with more than 100 layers, and Spikformer \cite{zhou2023spikformer} reaches 74.81\% top-1 accuracy on ImageNet at 66.3M parameters with 4 time steps, comparable in scale to second-generation ImageNet baselines. QKFormer \cite{zhou2024qkformer} extends this line further, reaching 85.65\% top-1 on ImageNet-1k at 64.96M parameters through spiking Q--K token attention, comparable to modest ANN Transformers of similar scale. Because the surrogate derivative has a weak biological motivation through smoothed or stochastic-threshold interpretations, but the optimization mechanism remains global, the learning grounding is \emph{Weak} rather than \emph{None} on the split-axis scheme of Section~\ref{sec:taxonomy_bio}.
 
Biologically observed local plasticity provides the contrasting case. Spike-timing-dependent plasticity, first demonstrated by Markram et al.\ \cite{markram1997regulation} and characterized by Bi and Poo \cite{bi1998synaptic} in cultured hippocampal neurons, updates synaptic strength based on the precise relative timing of pre- and post-synaptic spikes: presynaptic activation followed by postsynaptic spiking within roughly 20 ms induces long-term potentiation, the reverse order induces depression, and the rule depends only on local state at the synapse without a global error signal. Diehl and Cook \cite{diehl2015unsupervised} applied \gls{stdp} to MNIST digit recognition in an unsupervised \gls{lif} network with lateral inhibition and adaptive thresholds, reaching 95\% accuracy with 6,400 excitatory neurons. Under current evidence, biologically local spiking learning rules have not yet demonstrated the same scaling profile, in combined scale, task breadth, and accuracy, as surrogate-gradient or backpropagation-trained systems. \gls{stdp}-trained \glspl{snn} therefore sit at \emph{Mechanistic plasticity} on the learning axis, the opposite tier from surrogate-gradient \glspl{snn} on the same forward architecture.
 
\subsection{Section Synthesis}
\label{sec:spiking_synthesis}
 
The hybrid event-driven column achieves the highest forward biological grounding in the survey (\emph{Mechanistic plausibility}) but exhibits the widest split on the learning axis: the same \gls{lif} architecture appears with \emph{Weak} learning grounding under surrogate gradients and \emph{Mechanistic plasticity} under \gls{stdp}.
 
Biologically, the computational problem these families target is temporally extended sensorimotor and cognitive computation under tight communication constraints, addressed by cortex through sparse spike events, continuous membrane integration, local plasticity, and timing-sensitive synapses. \glspl{snn} preserve spikes, thresholds, membrane state, and, in the \gls{stdp} variant, local plasticity; they omit dendritic complexity, neuromodulation, multi-compartment computation, and the large-scale cortical learning organization that supports biological credit assignment. Section~\ref{sec:learning} takes up biologically plausible learning rules and Section~\ref{sec:neuromorphic} takes up the neuromorphic substrate.
 
\section{Biologically Plausible Learning and Credit Assignment}
\label{sec:learning}
 
Sections~\ref{sec:static} through~\ref{sec:spiking} traverse the taxonomy along the forward-dynamics axis. This section turns to the learning axis and asks the symmetric question: given the diversification of forward dynamics documented in the preceding sections, what alternatives to backpropagation exist for credit assignment, and do they close the forward--backward disconnect? Biologically plausible learning rules each relax one or more assumptions of backpropagation, but each relaxation usually trades away some combination of scalability, task performance, depth, or generality. The disconnect persists not because alternatives are absent but because no alternative currently matches backpropagation's combined scalability, generality, and hardware alignment.
 
\subsection{The Biological Credit-Assignment Problem}
\label{sec:learning_problem}
 
Backpropagation \cite{rumelhart1986learning} is the engineering baseline against which biologically plausible alternatives are measured, and its structural incompatibilities with cortical computation were articulated as early as Crick \cite{crick1989recent}. Whittington and Bogacz \cite{whittington2019theories} enumerate three canonical problems: the lack of a biologically local error representation, since each weight update depends on an error signal computed through downstream pathways rather than on variables directly available at the synapse; the symmetry of forward and backward weights, the weight-transport problem of \cite{crick1989recent}, since backpropagation transports error using the transpose of the forward weight matrix; and the unrealistic neuron models, since cortical neurons spike and the derivative of a spike is not well-defined. Three further constraints relevant to deep-learning practice complete the set of assumptions the alternatives in this section relax: update locking \cite{jaderberg2017decoupled}, since the backward pass cannot begin until the forward pass completes; stored activations, since the full forward state must be retained for the backward computation; and temporal unrolling \cite{werbos1990backpropagation}, since backpropagation through time requires the network state to be unrolled across the input sequence. Synaptic locality, in the sense that updates depend only on signals available at the synapse, is the binding biological constraint. The configurations surveyed below relax these constraints to varying degrees, summarized in Table~\ref{tab:relaxation_family}.
 
\subsection{Approximate-Gradient Alternatives}
\label{sec:learning_approx}
 
Feedback alignment \cite{lillicrap2016random} replaces the transpose of the forward weights in the backward pass with a fixed random matrix \(B\); forward weights adapt so that the pseudo-gradient aligns sufficiently with the true gradient to support learning, relaxing weight transport. The direct variant \cite{nokland2016direct} projects the output error to each hidden layer in parallel through independent fixed random matrices, additionally relaxing update locking. Target propagation \cite{lee2015difference} dispenses with gradient propagation altogether, assigning each layer a target activation propagated backward through learned approximate inverses. These relaxations do not preserve backpropagation's scaling: Bartunov et al.\ \cite{bartunov2018assessing} report target-propagation failure at ImageNet and feedback alignment reaching roughly one-quarter of BP top-1 accuracy on the same task, with the gap concentrated in convolutional and deep architectures. Methods that reintroduce weaker forms of weight symmetry close much of the gap. Weight Mirror and Kolen--Pollack learning \cite{akrout2019deep} adjust dynamic feedback weights and approach BP on ImageNet: ResNet-50 top-1 error is 23.4\% for Weight Mirror and 23.9\% for Kolen--Pollack versus 22.9\% for BP (within roughly one percentage point), substantially outperforming plain feedback alignment. Product Feedback Alignment \cite{li2024deep} reaches 68.46\% (PFA) and 69.30\% (PFA-o) top-1 accuracy on ResNet-18 ImageNet versus 69.69\% for BP, closing the gap to under half a percentage point without explicit weight symmetry. These results place the highest demonstrated scales of biologically-motivated learning in the family of error-broadcast approximate-gradient rules rather than in strictly local plasticity, and support the abstract's careful wording that the highest scales concentrate in global or closely gradient-derived and error-propagation mechanisms.
 
\subsection{Energy-Based and Inference-Based Learning}
\label{sec:learning_energy_inference}
 
Equilibrium propagation \cite{scellier2017equilibrium} casts learning as a contrastive update between two equilibria of an energy-based network: a free phase relaxes to \(\mathbf{s}^{\star}_{\mathrm{free}}\) minimizing \(E\), a nudged phase weakly clamping the output toward the target relaxes to \(\mathbf{s}^{\star}_{\beta}\) under \(F = E + \beta\ell\). Scellier and Bengio prove the resulting contrastive Hebbian update, using only local state, performs gradient descent equivalent to recurrent backpropagation in the infinitesimal-nudging limit. Symmetric \(\pm\beta\) nudging reduces estimator bias; Laborieux et al.\ \cite{laborieux2021scaling} reduced CIFAR-10 test error to \(11.7\%\), with extensions approaching backpropagation on downsampled ImageNet. Costs are the iterative equilibrium-settling cost and the strict energy-based architectural restriction.
 
Predictive coding \cite{rao1999predictive} provides the complementary mechanism: each layer maintains activity and prediction-error neurons; error neurons compute residuals \(\epsilon_i = (x_i - \mu_i)/\Sigma_i\) using the top-down prediction \(\mu_i\), and synaptic updates depend only on presynaptic activity and postsynaptic error. Whittington and Bogacz \cite{whittington2017approximation} prove this local Hebbian update produces gradients equivalent to backpropagation under unit variances and output-layer-dominant limit assumptions; Millidge et al.\ \cite{millidge2022predictive} extend the equivalence to arbitrary computation graphs but under a fixed-prediction assumption that further constrains biological plausibility.
 
Contrastive divergence \cite{hinton2002training} trains restricted Boltzmann machines through short Gibbs sampling chains, historically central to deep generative learning before end-to-end gradient methods became dominant. The learning grounding for this subsection ranges from \emph{Approximate or implicit gradient} (equilibrium propagation, predictive coding under the equivalence regime) to \emph{Local/algorithmic} (contrastive divergence).
 
\subsection{Local Plasticity Beyond Spiking STDP}
\label{sec:learning_local}
 
Spike-timing-dependent plasticity was treated in Section~\ref{sec:spiking} as the local-plasticity configuration of spiking networks; broader local-plasticity mechanisms beyond pairwise spike timing also span this row. Hebbian learning \cite{hebb1949organization} provides the elementary local rule \(\Delta w_{ij} \propto x_i x_j\), with Oja's rule and related stabilization mechanisms adding multiplicative decay to bound the weights. Three-factor learning rules \cite{fremaux2016neuromodulated} generalize Hebbian learning by gating the synaptic update with a global neuromodulatory signal and a synaptic eligibility trace, enabling reward- and error-modulated learning without a backward-propagated gradient. Miconi \cite{miconi2017biologically} instantiates this scheme in continuous-time recurrent networks, showing that reward-modulated Hebbian plasticity can reproduce neural dynamics observed in cognitive tasks. More recently, Dendritic Localized Learning \cite{lv2025dendritic} extends the direction to modern architectures with a spatially and temporally local, dendrite-inspired plasticity rule that trains ANNs, CNNs, ResNets, and Transformers competitively on CIFAR-10, CIFAR-100, and small ImageNet subsets, addressing multiple biological objections to backpropagation simultaneously, though not yet at full ImageNet-1k scale.
 
Hierarchical temporal memory \cite{hawkins2016neurons,hawkins2019framework} is the framework-level extreme of local-plasticity learning: a cortical-column-inspired architecture using sparse distributed binary representations and dendrite-local Hebbian-style learning, with temporal sequence memory provided by context-dependent activation rather than recurrent gradient flow. It is not a learning rule attached to a standard architecture but a framework with its own forward dynamics and local learning rules, sitting at \emph{Local or algorithmic} on the learning axis: the permanence-update rule is Hebbian-like in form but is an algorithmic abstraction rather than a directly measured plasticity mechanism.
 
E-prop \cite{bellec2020solution} is the strongest counterexample in the temporal--local corner. Applied to recurrent networks of leaky integrate-and-fire neurons with spike-frequency adaptation, e-prop replaces the offline unrolling of \gls{bptt} with per-synapse eligibility traces that maintain, at each synapse, a running mechanistic memory of pre--post coincidence; a neuron-specific broadcast learning signal from the output is then combined with the trace to produce a weight update online. The eligibility-trace component is synapse-local and removes backward propagation through time, while the update still depends on a neuron-specific broadcast learning signal; the important achievement is online learning without \gls{bptt}, not complete spatial locality. Bellec et al.\ demonstrate e-prop on TIMIT phoneme recognition, a delayed evidence-accumulation task modeled after rodent linear-track experiments, and Atari reinforcement learning (Pong and Fishing Derby), in each case narrowing but not closing the gap to \gls{bptt}. The demonstrated scale remains below full ImageNet-1k and GPT-class benchmarks, and no e-prop-trained recurrent network has yet matched a mainstream Transformer on language modeling. The instructive lesson is precisely this decomposition: eligibility traces are the biologically identified mechanism for local temporal credit assignment, while the residual dependence on a top-down broadcast identifies the coordinate on which further work is needed.
 
Synthetic gradients and decoupled neural interfaces \cite{jaderberg2017decoupled} target update locking, one of the six backpropagation assumptions of Section~\ref{sec:learning_problem}, rather than weight transport or global error. A small auxiliary network at each layer predicts the downstream gradient from the layer's own activations, allowing modules to update asynchronously without waiting for a full backward pass; the synthetic-gradient predictor is itself trained by minimizing the distance to the true gradient once it becomes available. Under the strict biological criterion of Section~\ref{sec:learning_problem} the mechanism is only weakly plausible, because each auxiliary predictor is itself a small backpropagation-trained network, but it is the operative example of a module-local, asynchronous training procedure and the reference construction for later work on decoupled and pipelined training. Jaderberg et al.\ demonstrated the method on feedforward and small recurrent networks; the original paper does not establish a general depth-dependent bias law, and reproducible scale-up to modern deep architectures remains limited.
 
\begin{table*}[!htbp]
\centering
\caption{Backpropagation assumptions relaxed by each biologically plausible learning family. R = relaxed; P = partially relaxed; \textendash{} = preserved. \emph{Global error} is marked P for energy- and inference-based methods where an output-layer error drives the relaxation but is not globally broadcast. \emph{Temporal unrolling} is preserved trivially for contrastive divergence since restricted Boltzmann machines have no temporal dimension.}
\label{tab:relaxation_family}
\renewcommand{\arraystretch}{0.85}
\setlength{\tabcolsep}{4pt}
\scriptsize
\begin{adjustbox}{max width=\textwidth}
\begin{tabular}{@{}lcccccc p{3.4cm}@{}}
\toprule
Family & \makecell{Weight\\transport} & \makecell{Global\\error} & \makecell{Update\\locking} & \makecell{Exact\\gradient} & \makecell{Stored\\activations} & \makecell{Temporal\\unrolling} & Main cost \\
\midrule
Feedback alignment \cite{lillicrap2016random} & R & \textendash & \textendash & P & \textendash & \textendash & Degrades on deep ImageNet \\
Direct feedback alignment \cite{nokland2016direct} & R & \textendash & R & P & \textendash & \textendash & Conv/deep accuracy gap \\
Target propagation \cite{lee2015difference} & R & P & P & R & \textendash & \textendash & Fails on ImageNet \\
Equilibrium propagation \cite{scellier2017equilibrium} & R & P & R & P & R & R & Equilibrium settling cost \\
Predictive coding \cite{whittington2017approximation} & R & P & P & P & R & P & Iterative relaxation cost \\
Contrastive divergence \cite{hinton2002training} & R & R & R & R & R & \textendash & Gibbs sampling cost \\
Three-factor / eligibility \cite{fremaux2016neuromodulated} & R & P & R & R & R & P & No deep credit assignment \\
e-prop \cite{bellec2020solution} & R & P & R & P & R & R & Small-to-mid-scale demonstrations \\
Synthetic gradients \cite{jaderberg2017decoupled} & \textendash & \textendash & R & P & R & \textendash & Limited deep-scale evidence \\
HTM / dendrite-local \cite{hawkins2016neurons} & R & R & R & R & R & R & Small-scale demonstrations \\
\bottomrule
\end{tabular}
\end{adjustbox}
\end{table*}
 
\subsection{Section Synthesis}
\label{sec:learning_synthesis}
 
The alternatives divide into two kinds. Learning-rule replacements target an existing forward architecture: feedback alignment, target propagation, synthetic gradients, equilibrium propagation, predictive coding, and e-prop each modify only the backward mechanism, preserving compatibility with mainstream practice while relaxing biological constraints only partially. Frameworks whose learning rule is tied to an alternative forward model include hierarchical temporal memory (local-plasticity) and restricted Boltzmann machines (energy-based); these strengthen biological grounding but have not matched the scale or task breadth of global-gradient systems. Table~\ref{tab:relaxation_family} makes the pattern explicit: no family relaxes the full set of backpropagation assumptions while preserving deep-learning scalability and generality. Section~\ref{sec:neuromorphic} turns to the substrate question, where the algorithmic gap documented here has a material counterpart.
 
\section{Neuromorphic Hardware}
\label{sec:neuromorphic}
 
The preceding sections established the forward--backward disconnect at the algorithmic level: forward dynamics have diversified along the taxonomy of Sections~\ref{sec:static} through~\ref{sec:spiking}, but the learning-side alternatives surveyed in Section~\ref{sec:learning} have not closed the gap. The compute substrate determines which learning rules and state dynamics are computationally natural. Neuromorphic hardware changes the substrate contract relative to \gls{gpu}/\gls{tpu} fabrics but cannot by itself supply a scalable, biologically plausible credit-assignment rule.
 
\subsection{The Hardware Lottery and Dense Gradient Substrates}
\label{sec:neuromorphic_lottery}
 
The hardware lottery of Hooker \cite{hooker2021hardware} holds that a research idea's success depends as much on the hardware available to test it as on its merit; applied to the algorithmic disconnect, this becomes a substrate-level claim that backpropagation is dominant in part because it is hardware-compatible. The dense-multiply-accumulate substrate that mainstream deep learning runs on, exemplified by the systolic-array architecture of the tensor processing unit \cite{jouppi2017indatacenter}, is purpose-built for matrix multiplication at high arithmetic intensity. The match with backpropagation is multidimensional: dense matmul is hardware-optimal, batch-parallel dataflow supplies global synchronization, high memory bandwidth makes stored activations tolerable, and batched training amortizes per-sample overhead. Biologically plausible learning rules typically achieve poorer utilization on the same substrate, since sparse asynchronous events, per-unit local state, and online updates map less favorably onto its dataflow, cache hierarchy, and enforced memory--arithmetic separation.
 
\subsection{Neuromorphic Computing Principles}
\label{sec:neuromorphic_principles}
 
Neuromorphic computing, introduced by Mead \cite{mead1990neuromorphic}, imposes a different computational contract from the von Neumann model on which \glspl{gpu} are built. The defining property \cite{indiveri2015memory} is co-location of memory and processing: synaptic state lives near or inside the computing element. Recent reviews \cite{schuman2022opportunities,roy2019towards} consolidate the operating principles: event-driven rather than clock-synchronous arithmetic; sparse, asynchronous, binary spike-based communication; per-neuron and per-synapse local state; online and continual operation without separate training and inference phases; and algorithm--hardware co-design. Representative platforms span digital--analog and on-chip-learning axes: \emph{TrueNorth} \cite{merolla2014million} (fully digital, no on-chip learning), \emph{Loihi} \cite{davies2018loihi} and \emph{Loihi 2} \cite{orchard2021loihi2} (programmable on-chip plasticity; Loihi~2 adds stateful spiking neurons), \emph{SpiNNaker} \cite{furber2014spinnaker} (massively parallel ARM cores optimized for spike routing), \emph{BrainScaleS-2} \cite{pehle2022brainscales2} (mixed-signal, hybrid on-chip plasticity).
 
\subsection{Substrate Alignment and Remaining Bottlenecks}
\label{sec:neuromorphic_bottlenecks}
 
\begin{figure*}[!htbp]
\centering
\begin{adjustbox}{max width=\textwidth}
\begin{tikzpicture}[
  >=Latex,
  every node/.style={font=\small},
  substbox/.style={
    rectangle, draw=black!75, thick, rounded corners=3pt,
    minimum width=6.6cm, minimum height=2.5cm,
    inner sep=6pt,
    align=center,
  },
  outcomebox/.style={
    rectangle, draw=black!60, thick, rounded corners=2pt,
    fill=blue!8,
    minimum width=5.6cm, minimum height=0.9cm,
    align=center,
  },
  bridgebox/.style={
    rectangle, draw=red!70, very thick, dashed, rounded corners=3pt,
    fill=red!4,
    minimum width=9.6cm, minimum height=1.35cm,
    align=center,
    inner sep=7pt,
  },
  flow/.style={-Latex, thick, black!80},
  workaround/.style={-Latex, thick, dashed, gray!75},
  missing/.style={-Latex, thick, dashed, red!70},
  pathlabel/.style={font=\itshape\bfseries\small, color=black!75},
]
 
\node[pathlabel] at (-5.5, 4.95) {Dense-gradient path};
\node[pathlabel] at (5.5, 4.95) {Event-driven path};
 
\node[substbox] (gpu) at (-5.5, 3.2) {%
  \textbf{\sffamily GPU / TPU substrate}\\[2pt]
  \footnotesize
  Dense tensors\\
  Stored activations\\
  Batch synchronization\\
  Global gradient transport
};
 
\node[substbox] (neuro) at (5.5, 3.2) {%
  \textbf{\sffamily Neuromorphic substrate}\\[2pt]
  \footnotesize
  Sparse events\\
  Local state\\
  Online updates\\
  Memory near compute
};
 
\node[outcomebox] (bp) at (-5.5, 0.8) {Backpropagation training scales};
\node[outcomebox] (ev) at (5.5, 0.8) {Event-driven inference scales};
 
\draw[flow] (gpu.south) -- (bp.north)
  node[midway, right=2pt, font=\scriptsize\itshape] {global gradient};
\draw[flow] (neuro.south) -- (ev.north)
  node[midway, right=2pt, font=\scriptsize\itshape] {sparse, local};
 
\draw[workaround] (bp.east) -- (ev.west)
  node[midway, above=3pt, font=\scriptsize, align=center] {off-chip surrogate-gradient training,\\on-chip event-driven deployment};
 
\node[bridgebox] (bridge) at (0, -1.9) {%
  \textbf{\color{red!70!black} Missing bridge}\\[2pt]
  \footnotesize Scalable local credit assignment\\
  (no substrate-aligned learning rule surveyed in Section~\ref{sec:learning} closes this gap)
};
 
\draw[missing] (bp.south) to[bend right=8] (bridge.north west);
\draw[missing] (ev.south) to[bend left=8] (bridge.north east);
 
\end{tikzpicture}
\end{adjustbox}
\caption{Substrate reconnection map. The \emph{dense-gradient path} on the left (\gls{gpu}/\gls{tpu}; backpropagation training scales) and the \emph{event-driven path} on the right (neuromorphic; event-driven inference scales) currently bridge only empirically (dashed gray): training runs off-chip with surrogate gradients on \glspl{gpu} and deploys on neuromorphic hardware for inference. The missing bridge (dashed red) is a substrate-aligned learning rule competitive with backpropagation at deep-network scale, which Section~\ref{sec:learning} found absent. Neuromorphic hardware removes the substrate penalty for event-driven forward computation but does not supply the missing rule.}
\label{fig:substrate_reconnection}
\Description{A conceptual schematic in three vertical layers. Top layer: two rounded rectangles side by side. The left box, labeled GPU/TPU substrate, lists dense tensors, stored activations, batch synchronization, and global gradient transport. The right box, labeled Neuromorphic substrate, lists sparse events, local state, online updates, and memory near compute. Middle layer: two outcome boxes shaded light blue. The left says Backpropagation training scales; the right says Event-driven inference scales. Solid black arrows connect each substrate down to its corresponding outcome, labeled global gradient on the left and sparse, local on the right. A dashed gray arrow runs horizontally between the two outcome boxes labeled off-chip surrogate-gradient training, on-chip event-driven deployment, indicating the current empirical workaround. Bottom layer: a dashed red rectangle labeled Missing bridge, with subtext scalable local credit assignment and a note that no substrate-aligned learning rule surveyed in Section 9 closes this gap. Dashed red arrows from each outcome box curve down toward this missing-bridge box, visually marking the unresolved connection.}
\end{figure*}
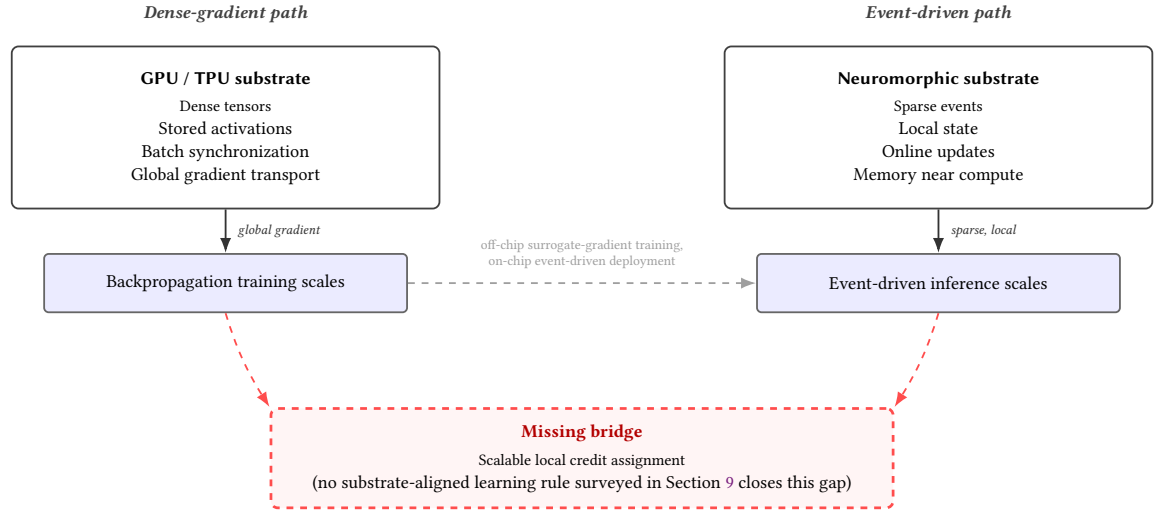

Figure~\ref{fig:substrate_reconnection} maps the two substrates side by side as the operating contracts they impose. The left column lists the structural primitives that the \gls{gpu}/\gls{tpu} substrate handles natively, and the right column lists those that the neuromorphic substrate handles natively. Each column delivers a substrate-aligned outcome: backpropagation training scales on the left, event-driven inference scales on the right. The current bridge between the two is empirical and asymmetric: large spiking systems are typically trained off-chip on \glspl{gpu} with surrogate gradients and deployed on neuromorphic hardware for inference \cite{eshraghian2023training}. The missing bridge is the converse, a substrate-aligned learning rule competitive with backpropagation at deep-network scale, which Section~\ref{sec:learning} found absent.
 
The forward-side gains are well documented. The original Loihi report \cite{davies2018loihi} demonstrates more than three orders of magnitude better energy-delay product than conventional solvers on LASSO sparse-coding workloads; the Loihi survey \cite{davies2021advancing} reports up to five orders of magnitude advantage in time-to-solution and six in energy consumption on the largest single-chip locally competitive sparse-coding workload (around \(10^5\) unknowns) relative to a CPU FISTA implementation. Local state dynamics map onto per-neuron state variables, online plasticity is implementable natively, and event-driven sensor processing integrates without conversion overhead.
 
The learning-side limitations are not removed by the change of substrate. Deep credit assignment has no substrate-level mechanism beyond the local plasticity rules of Section~\ref{sec:learning}; global objectives still require either backpropagation (unnatural on neuromorphic substrates) or an alternative learning rule with the scalability properties Section~\ref{sec:learning} found absent. Training of large \glspl{snn} consequently runs off-chip on \glspl{gpu} through surrogate-gradient backpropagation \cite{eshraghian2023training}, so substrate alignment applies to inference rather than training. Conversion of backpropagation-trained networks to spike-encoded inference is feasible: early methods needed long latency windows (order two-to-three thousand time steps for ImageNet-class inference \cite{sengupta2019going}), but recent conversion methods have narrowed this gap substantially, with SpikeZIP-TF \cite{you2024spikezip} achieving 83.82\% ImageNet-1k top-1 accuracy through exact ANN-to-SNN equivalence at approximately 8 time steps for CNN structures and comparable low-latency extensions to Transformer-based \glspl{snn}. The most recent authoritative assessment \cite{kudithipudi2025neuromorphic} flags four ecosystem bottlenecks: low-level software toolchains, hardware-dependent algorithms, absent community consensus on performance metrics, and largely absent formal complexity arguments for neuromorphic advantage.
 
\subsection{Section Synthesis}
\label{sec:neuromorphic_synthesis}
 
The substrate contract is asymmetric and must be stated carefully: neuromorphic energy efficiency is real but workload-specific, conditioned on sparsity; on-chip learning is supported on Loihi and BrainScaleS-2 \cite{davies2018loihi,pehle2022brainscales2} but is restricted to local plasticity variants rather than backpropagation equivalents. Hardware removes the substrate penalty for local and event-driven computation but does not by itself produce a scalable global credit-assignment rule. Section~\ref{sec:open_problems} treats dynamics, learning rule, and substrate as three coupled levers.
 
\section{Cross-Family Synthesis and Open Problems}
\label{sec:open_problems}
 
Read across the traversal, the taxonomy exposes structural patterns and open problems not visible from any single family.
 
\subsection{What the Taxonomy Reveals}
\label{sec:open_synthesis}
 
Figure~\ref{fig:master_taxonomy}, Table~\ref{tab:model_families}, and the count matrix in Table~\ref{tab:config_count_matrix} expose three patterns that no single family makes visible.
 
\emph{First}, forward dynamics diversify across the five mathematically distinct classes defined by the taxonomy: Static, Discrete-time/sequence, Continuous-time, Implicit, and Hybrid event-driven. Attention and structured state-space models populate the Discrete-time/sequence class, while modern Hopfield retrieval and \glspl{deq} illustrate distinct forms of the Implicit class. Every forward class in Figure~\ref{fig:master_taxonomy} is populated.
 
\emph{Second}, the learning side has a different distribution. Categorically, Global gradient is the largest single row (16 of 32), followed by Approximate or implicit gradient (7 of 32). The mechanism-level decomposition of Supplementary Section~S3 confirms the concentration under a finer partition: ordinary reverse-mode backpropagation alone accounts for ten configurations, while the Approximate/implicit category fragments across seven individually represented mechanisms. Independently, the scaling evidence of Section~\ref{sec:learning} and Supplementary Section~S4 shows that the highest demonstrated scales remain concentrated in global or closely gradient-derived and error-propagation mechanisms, even after the recent Weight Mirror, Kolen--Pollack, Product Feedback Alignment, e-prop, and Dendritic Localized Learning results considered in this survey.
 
\emph{Third}, biological grounding is not inherited automatically from the forward architecture. Surrogate-gradient and \gls{stdp}-trained spiking networks share the same \gls{lif} forward operator at \emph{Mechanistic plausibility} on the forward axis but occupy opposite tiers on the learning axis (\emph{Weak} versus \emph{Mechanistic plasticity}). The split-axis scheme of Section~\ref{sec:taxonomy_bio} is what makes this independence visible: forward grounding and learning grounding move separately, and the same architecture--learning configuration is the correct unit of analysis rather than the architecture name alone.
 
\subsection{Credit Assignment Beyond Backpropagation}
\label{sec:open_credit}
 
The central open problem is a scalable local credit-assignment rule. No alternative surveyed in Section~\ref{sec:learning} relaxes the six assumptions of Section~\ref{sec:learning_problem} while preserving scalability, task breadth, and hardware compatibility; Table~\ref{tab:relaxation_family} shows every family pays a cost in scale, depth, or generality. Credit assignment has two sub-problems: \emph{spatial} (across depth at a single time) and \emph{temporal} (across time). \gls{bptt} reduces temporal to spatial by unrolling time, and Section~\ref{sec:learning} relaxations address mostly the spatial sub-problem; the temporal sub-problem is where the gap is widest. The scaling component is empirically falsifiable: a strictly local or mechanistic-plasticity rule reaching backpropagation-parity performance at full ImageNet-1k or GPT-class language-model scale under comparable training compute would revise it (Supplementary Section~S4 states the operational criterion).
 
\subsection{Dynamics, Time, and Biological Structure}
\label{sec:open_dynamics_biology}
 
Recurrent, continuous-time, and spiking models make time explicit in the forward pass, but their credit-assignment machinery (\gls{bptt}, adjoint methods, surrogate-gradient training) all remain globally synchronized procedures that unroll or solver-trace time before gradients flow. Eligibility traces in three-factor learning \cite{fremaux2016neuromodulated} and e-prop \cite{bellec2020solution} furnish a biologically plausible mechanism (local transient memory holding pre-post coincidence until a modulatory signal arrives), and e-prop attacks temporal credit assignment in recurrent spiking networks without \gls{bptt} at scale below deep-learning benchmarks. Forward computation is also limited at the single-neuron level: dendritic structure \cite{poirazi2003pyramidal,larkum2013cellular,beniaguev2021single} is a candidate substrate for compartment-local teaching signals without a global backward broadcast.
 
\subsection{Benchmarking and a Research Agenda for Configuration-Level Co-Design}
\label{sec:open_codesign}
 
Current evaluation conventions emphasize task accuracy and parameter scale but rarely separate state dynamics, learning locality, temporal credit assignment, and substrate cost. Cross-family comparisons should therefore report these dimensions jointly rather than treating accuracy on dense hardware as sufficient evidence of configuration-level scalability.
 
The synthesis above implies four research priorities that would together address combinations and capabilities that remain underrepresented across the taxonomy and substrate analysis: (i) a scalable local credit-assignment rule that addresses both spatial and temporal sub-problems without unrolling time into space; (ii) learning mechanisms that exploit dendritic and compartmental structure, providing compartment-specific teaching signals without global backward broadcast; (iii) substrate-aligned on-chip deep learning that trains, and not only deploys, non-trivial models on neuromorphic hardware; and (iv) configuration-level benchmarks that jointly report accuracy, locality, temporal behavior, memory, and energy cost so that architecture--learning--substrate combinations can be compared against one another rather than only against dense-substrate baselines. These are the research implications of the survey rather than observed findings; each names a joint state-dynamics/credit-assignment/substrate combination for which the taxonomy exposes a gap. The sparse-region audit in Supplementary Section~S5 further cautions against interpreting every empty cell as a research gap: three of four initially flagged combinations were occupied after configuration-level reclassification, while continuous-time non-equilibrium dynamics paired with energy or unsupervised learning remained near-empty.
 
\section{Conclusion}
\label{sec:conclusion}
 
Forward neural computation diversifies across five state-dynamics classes: Static, Discrete-time/sequence, Continuous-time, Implicit, and Hybrid event-driven. Scalable credit assignment does not distribute in the same way. Figure~\ref{fig:master_taxonomy} shows five populated forward classes feeding a more concentrated backward distribution: Global gradient is the largest single category (16 of 32 configurations), Approximate or implicit gradient the second (7 of 32). The finer mechanism-level decomposition of Supplementary Section~S3 confirms the concentration: ordinary reverse-mode backpropagation alone accounts for ten configurations, while the Approximate/implicit category fragments across seven individually represented mechanisms. Figure~\ref{fig:timeline} provides the complementary historical orientation.
 
The joint taxonomy organizes neural computation along coupled state-dynamics and credit-assignment axes, with biological grounding assigned independently to the forward operator and the learning rule. Its atomic unit is the architecture--learning configuration rather than the architecture name in isolation: the same forward dynamics occupies different taxonomic positions depending on how credit is assigned. Biology functions as a diagnostic constraint, identifying which mechanisms a configuration preserves, which it discards, and which engineering substitutes compensate for the omissions. Surrogate-gradient and STDP-trained spiking networks share the same strongly grounded forward operator but occupy opposite tiers on the learning axis.
 
Alternatives to backpropagation exist. Recent results span error-broadcast approximate-gradient mechanisms (Weight Mirror, Kolen--Pollack, Product Feedback Alignment) and more local alternatives based on eligibility traces or dendritic plasticity (e-prop, Dendritic Localized Learning), but no single configuration yet combines locality, temporal credit assignment, task breadth, scalability, and substrate alignment. Hardware alignment removes the substrate penalty for sparse event-driven computation without supplying a scalable learning rule. Closing the disconnect therefore likely requires configuration-level co-design in which state dynamics, credit assignment, and computational substrate are treated as a joint design object rather than optimized independently. Whether such configurations can match the scalability of contemporary global-gradient systems remains an open empirical question.



\bibliographystyle{ACM-Reference-Format}
\bibliography{references}

\end{document}


\maketitle
\thispagestyle{fancy}

\noindent
\textbf{Status of this document.} This document is the electronic supplement to the main manuscript. It contains methodological detail (literature search procedure, adjudication rationales for ambiguous configurations, and the per-configuration ledger that generates Figure~2 of the main text) together with two evidence audits supporting the survey's scaling claims.

\noindent
\textbf{Content.} This appendix documents the literature search procedure, the placement decisions for architecture--learning configurations whose taxonomy placement required adjudication, the per-configuration ledger reconciling Figure~2 of the main text, the empirical-evidence base underlying the scaling claim of Section~\ref{sec:supp_scaling}, and the audit of apparently sparse regions of the joint taxonomy. Section references prefixed with ``S'' refer to this document; unprefixed section references refer to the main manuscript.

\section{Literature Search Procedure}
\label{sec:supp_litsearch}

The procedure summarized in Section~3.5 of the main manuscript is documented below in its operational form: the databases queried, the temporal windows of the search, the Boolean search strings grouped by axis of the taxonomy, the inclusion and exclusion criteria, the adjudication rule for ambiguous classifications, and the limitations the procedure imposes on the survey's coverage.

\subsection{Databases}
\label{sec:supp_litsearch_databases}

\begin{itemize}[leftmargin=*, itemsep=2pt]
  \item ACM Digital Library
  \item IEEE Xplore
  \item Scopus
  \item arXiv
  \item Google Scholar (used primarily for citation chaining from results obtained in the preceding databases)
\end{itemize}

\subsection{Search windows}
\label{sec:supp_litsearch_windows}

Three overlapping windows were used:
\begin{itemize}[leftmargin=*, itemsep=2pt]
  \item \textbf{Foundational:} no lower bound; works that established a model family or learning rule covered by the taxonomy.
  \item \textbf{Representative:} 2010--2025; canonical works that define a family's modern form.
  \item \textbf{Recent high-impact:} 2020--2026; developments that have reshaped the field's current trajectory.
\end{itemize}

\subsection{Search strings}
\label{sec:supp_litsearch_strings}

Boolean search strings, grouped by the axis of the taxonomy they target. Strings were applied with appropriate field restrictions (title, abstract, keywords) on each database. Synonyms within parentheses were used as disjunctions.

\begin{itemize}[leftmargin=*, itemsep=4pt]
  \item \textbf{State-dynamics terms.} (``recurrent neural network'' OR ``hidden state'') AND (``state-space model'' OR ``selective state-space'' OR ``structured state-space''); (``neural ordinary differential equation'' OR ``Neural ODE'' OR ``continuous-time neural network'' OR ``liquid time-constant''); (``deep equilibrium model'' OR ``fixed-point network'' OR ``implicit layer''); (``Hopfield network'' OR ``dense associative memory'' OR ``modern Hopfield''); (``spiking neural network'' OR ``leaky integrate-and-fire'' OR ``LIF neuron'' OR ``event-driven neural'').
  \item \textbf{Credit-assignment and learning-rule terms.} (``backpropagation'' OR ``backpropagation through time'' OR ``BPTT''); (``feedback alignment'' OR ``direct feedback alignment'' OR ``sign-symmetry'' OR ``target propagation''); (``weight mirror'' OR ``Kolen-Pollack'' OR ``product feedback alignment''); (``equilibrium propagation''); (``predictive coding'' AND (``error backpropagation'' OR ``local Hebbian'' OR ``free-energy'' OR ``prospective configuration'')); (``Forward-Forward'' OR ``forward-forward algorithm''); (``surrogate gradient'' AND ``spiking''); (``e-prop'' OR ``eligibility propagation''); (``synthetic gradient'' OR ``decoupled neural interface''); (``three-factor learning rule'' OR ``reward-modulated Hebbian'' OR ``eligibility trace'').
  \item \textbf{Biological-plausibility terms.} (``biologically plausible'' OR ``biologically constrained'' OR ``brain-inspired'') AND (``credit assignment'' OR ``learning rule'' OR ``synaptic plasticity''); (``spike-timing-dependent plasticity'' OR ``STDP''); (``dendritic computation'' OR ``compartmental neuron''); (``neuromodulation'' OR ``three-factor plasticity''); (``cortical microcircuit'' AND (``predictive coding'' OR ``hierarchical inference'')).
  \item \textbf{Neuromorphic and hardware terms.} (``neuromorphic computing'' OR ``neuromorphic processor'' OR ``neuromorphic hardware''); (``Loihi'' OR ``TrueNorth'' OR ``SpiNNaker'' OR ``BrainScaleS'' OR ``Akida'' OR ``Tianjic''); (``on-chip learning'' OR ``on-chip plasticity'' OR ``in-memory computing'' AND ``synaptic''); (``hardware lottery'' OR ``algorithm-hardware co-design'').
\end{itemize}

\subsection{Last-search date}

Last comprehensive search: June 2026. Citation chaining and venue-publication-status verification (for arXiv preprints, conference papers awaiting final venue, and recent 2024--2026 publications) was repeated immediately before manuscript submission.

\subsection{Inclusion criteria}
\label{sec:supp_litsearch_include}

A work is included in the survey when it satisfies all of the following:
\begin{enumerate}[leftmargin=*, itemsep=2pt]
  \item It meets one of the two scope criteria stated in Section~3.5 of the main manuscript: it instantiates a nontrivial form of state dynamics in the forward computation, or it is substantively grounded in neuroscience at the architectural or learning-rule level.
  \item It admits classification under the formal definitions of state-dynamics structure (main manuscript Section~3.1), credit-assignment mechanism (main manuscript Section~3.2), and biological grounding (main manuscript Section~3.3).
  \item It either defines or instantiates an architecture--learning configuration in the sense of main manuscript Section~3.2 (foundational paragraph).
\end{enumerate}

\subsection{Exclusion criteria}
\label{sec:supp_litsearch_exclude}

A work is excluded when it falls outside the joint state-dynamics and credit-assignment axes; the model families excluded on this basis are listed in Section~3.5 of the main manuscript, with the rationale for each exclusion.

\subsection{Classification adjudication}
\label{sec:supp_litsearch_adjudicate}

The formal definitions in Sections~3.1--3.3 of the main manuscript are authoritative for classification. Configurations whose initial placement was ambiguous are documented in Section~\ref{sec:supp_audit} below, with the rationale for the final placement.

\subsection{Limitations of the procedure}
\label{sec:supp_litsearch_limits}

The procedure carries the standard limitations of database-driven survey work:
\begin{itemize}[leftmargin=*, itemsep=2pt]
  \item \emph{Language and indexing bias.} English-language works indexed in the databases listed above are over-represented.
  \item \emph{Taxonomy-induced selection bias.} Works that do not fit the two-axis taxonomy are under-represented or excluded, even when they may be substantively relevant to neural computation. This is a deliberate consequence of the scope choice.
  \item \emph{Recency bias.} The high-impact window prioritizes 2020--2026 developments, which may overstate the importance of recent work relative to its eventual standing.
\end{itemize}

\subsection{Preprint policy}
\label{sec:supp_preprints}

arXiv and bioRxiv preprints are included on the same footing as peer-reviewed publications when they either (i) introduce or first characterize a model family that later became a stable object of study in the taxonomy, or (ii) report the strongest available scaling evidence for a family within the search window and had no subsequent peer-reviewed superseding version at the cutoff. Preprint entries are cited with the arXiv identifier and year; the survey does not treat preprint status as a quality gate but does flag preprint-only evidence explicitly in the accompanying prose whenever it anchors a scaling or performance claim.

\subsection{Family selection}
\label{sec:supp_family_selection}

Model families included in the model-family table in Section~3.3 of the main manuscript are selected by three complementary criteria: (i) representativeness of one or more state-dynamics classes defined in Section~3.1, (ii) representativeness of one or more credit-assignment classes defined in Section~3.2, and (iii) sufficient primary-literature coverage that a stable placement in the joint taxonomy can be defended from the sources rather than inferred. Emerging families with insufficient primary evidence to fix a placement are noted in the prose but not included as separate rows. Multiple representative papers per family are cited when needed to fix the placement.

\subsection{Configuration-counting rule}
\label{sec:supp_config_rule}

A counted configuration in Figure~2 of the main manuscript, and equivalently a row of the ledger in Section~\ref{sec:supp_ledger}, requires a concrete architecture--learning pairing supported by primary evidence: the paper must define both the forward architecture (state-dynamics class) and the learning rule (credit-assignment class) used to train it, and the pair must fit the joint taxonomy without requiring a re-derivation of either component. Conceptual papers, equivalence proofs, and re-interpretations of existing training pipelines support classification of already-counted configurations but do not by themselves create new ledger rows; a paper showing that predictive coding recovers backpropagation under specific assumptions, for example, refines the placement of an existing predictive-coding configuration but does not add a new row unless it also demonstrates a distinct concrete pairing. Counts therefore describe the audited representative set that anchors the taxonomy rather than the prevalence of each combination in the literature; readers interested in prevalence should consult the individual sections of the main manuscript for cited representatives.

\section{Architecture--Learning Configuration Audit}
\label{sec:supp_audit}

Configurations whose initial taxonomy placement required adjudication are recorded below, each with its state-dynamics class (per Section~3.1 of the main manuscript), credit-assignment row (per Section~3.2), forward grounding tier (per Section~3.3), learning grounding tier (per Section~3.3), and the rationale that anchors the placement. Configurations whose primary classification remains contested are recorded with a dual reading: a primary commitment and a secondary alternative.

\subsection{Modern Hopfield retrieval in end-to-end trained networks}
\label{sec:supp_audit_hopfield_modern}

\noindent
\textbf{Reference.} Ramsauer et al., ``Hopfield Networks is All You Need,'' ICLR 2021.

\noindent
\textbf{Placement.}
\begin{itemize}[leftmargin=*, itemsep=2pt]
  \item State-dynamics class: Implicit / fixed-point (one-step retrieval in practice; the underlying dynamics admit a continuous-state energy with attractor fixed points).
  \item Credit-assignment row: Global gradient.
  \item Forward grounding: \textbf{Functional analogy} (primary). Historical inspiration is also defensible; architectural constraint is rejected under the strict criterion.
  \item Learning grounding: None.
\end{itemize}

\noindent
\textbf{Rationale.} The modern Hopfield update is mathematically equivalent to transformer self-attention under the inverse-temperature setting $\beta = 1/\sqrt{d_k}$ (verified against the original paper; the equivalence is exact and single-step). Because the resulting layer is identical to a mainstream design that was not adopted on biological grounds, biology does not rule out alternatives in the sense required by the strict architectural-constraint criterion of Section~3.3 of the main manuscript. The associative-memory framing is post-hoc relative to the modern continuous-state formulation; explicit biologically plausible variants appear later in the program (Krotov \& Hopfield, ``Large associative memory problem in neurobiology and machine learning,'' ICLR 2021), reinforcing that the original modern formulation's biological grounding is functional rather than constraint-derived.

\subsection{Classical Hopfield network with Hebbian one-shot storage}
\label{sec:supp_audit_hopfield_classical}

\noindent
\textbf{Reference.} Hopfield, ``Neural networks and physical systems with emergent collective computational abilities,'' PNAS 79(8):2554--2558, 1982.

\noindent
\textbf{Placement.}
\begin{itemize}[leftmargin=*, itemsep=2pt]
  \item State-dynamics class: Discrete-time recurrence with fixed-point function (asynchronous binary updates converging to attractor states).
  \item Credit-assignment row: Local plasticity.
  \item Forward grounding: Architectural constraint.
  \item Learning grounding: Local or algorithmic.
\end{itemize}

\noindent
\textbf{Rationale.} Storage is the outer-product Hebbian rule, satisfying the synapse-local update criterion. Symmetric recurrent connectivity and attractor architecture are derived from the collective biological account in the 1982 paper, and the symmetric-weight constraint is a substantive structural commitment rather than an engineering convenience. Critical storage load $\alpha_c \approx 0.138$ (Amit, Gutfreund \& Sompolinsky, Phys.\ Rev.\ Lett.\ 1985) is the capacity ceiling that motivates the modern dense formulations.

\subsection{Predictive coding: four variants}
\label{sec:supp_audit_pc}

Predictive coding is not a single configuration but a family with distinguishable variants that disagree on at least one taxonomy axis. Each variant is classified separately below.

\subsubsection{Recurrent-inference variant}
\label{sec:supp_audit_pc_recurrent}

\noindent
\textbf{Reference.} Rao \& Ballard, ``Predictive coding in the visual cortex,'' Nature Neuroscience 2(1):79--87, 1999.

\noindent
\textbf{Placement.} Continuous-time (primary; implicit / fixed-point reading available); Local plasticity; Architectural constraint; Local or algorithmic.

\noindent
\textbf{Rationale.} The model posits feedback connections carrying predictions of lower-level activity and feedforward connections carrying residual errors, an architecture motivated by cortical anatomy and explaining extra-classical receptive-field effects. Inference is a relaxation dynamic minimizing squared prediction error; learning adjusts the generative weight matrix via locally available errors. The forward architecture qualifies as architectural constraint because the predict-down/error-up wiring is dictated by the cortical hypothesis rather than chosen for performance.

\subsubsection{Energy-based (free-energy) variant}
\label{sec:supp_audit_pc_energy}

\noindent
\textbf{References.} Friston, ``A theory of cortical responses,'' Phil.\ Trans.\ R.\ Soc.\ B 360(1456):815--836, 2005; Friston, ``The free-energy principle: a unified brain theory?,'' Nature Reviews Neuroscience 11(2):127--138, 2010; Friston \& Kiebel, ``Predictive coding under the free-energy principle,'' Phil.\ Trans.\ R.\ Soc.\ B 364(1521):1211--1221, 2009.

\noindent
\textbf{Placement.} Continuous-time / implicit fixed-point; Energy or unsupervised; Architectural constraint; Local or algorithmic.

\noindent
\textbf{Rationale.} The free-energy formulation casts perception and learning as descent on a single variational free-energy functional, which is the canonical energy-based reading. The hierarchical message-passing microcircuit (prediction units, error units, precision weighting) maps onto biologically realistic cortical microcircuitry; precision corresponds to synaptic gain. The energy row is the most faithful for this variant because the objective is an explicit (free-)energy functional rather than an exact global gradient. Some authors treat this variant as functionally equivalent to the recurrent-inference variant above; the audit retains the distinction because the energy/variational formalism and the receptive-field instantiation make different commitments.

\subsubsection{Local-error (Hebbian-approximation) variant}
\label{sec:supp_audit_pc_localerror}

\noindent
\textbf{Reference.} Whittington \& Bogacz, ``An Approximation of the Error Backpropagation Algorithm in a Predictive Coding Network with Local Hebbian Synaptic Plasticity,'' Neural Computation 29(5):1229--1262, 2017.

\noindent
\textbf{Placement.} Implicit / fixed-point; Approximate or implicit gradient; Functional analogy; Local or algorithmic (with a mechanistic-plasticity reading available).

\noindent
\textbf{Rationale.} Value nodes relax to an equilibrium of the inference dynamics before weight updates, and for certain parameter regimes the resulting weight change converges to the backpropagation update. The stated aim is to show cortical networks with simple local Hebbian plasticity can approximate error backpropagation, performing supervised learning autonomously with synapse-local information. The forward architecture is a layered predictive-coding network motivated by cortical hierarchy but used here as a means to approximate backpropagation; this defeats the strict architectural-constraint test and places forward grounding at functional analogy.

\subsubsection{Backpropagation-equivalent limit variant}
\label{sec:supp_audit_pc_bpequiv}

\noindent
\textbf{References.} Millidge, Tschantz \& Buckley, ``Predictive Coding Approximates Backprop along Arbitrary Computation Graphs,'' Neural Computation 34(6):1329--1368, 2022; Song, Lukasiewicz, Xu \& Bogacz, ``Can the brain do backpropagation?---Exact implementation of backpropagation in predictive coding networks,'' NeurIPS 2020; Song, Millidge, Salvatori, Lukasiewicz, Xu \& Bogacz, ``Inferring neural activity before plasticity as a foundation for learning beyond backpropagation,'' Nature Neuroscience 27(2):348--358, 2024.

\noindent
\textbf{Placement.} Implicit / fixed-point; Approximate or implicit gradient; Functional analogy; Local or algorithmic.

\noindent
\textbf{Rationale (contested classification, decided).} The defining property of this variant is gradient equivalence to backpropagation along arbitrary computation graphs, demonstrated on SVHN/CIFAR-10/CIFAR-100 at roughly two orders of magnitude greater compute than direct backpropagation. The local-update structure (each weight update uses local prediction errors) earns the local-algorithmic learning grounding, but mechanistic plausibility is not warranted because the architecture is a generic computation graph chosen to mirror an ANN rather than a biologically dictated structure. Song et al.\ 2024 strengthen the neuroscience interpretation through the prospective-configuration property, but this refines the rationale of the local-algorithmic tier rather than upgrading the classification to mechanistic plasticity.

\subsection{Equilibrium propagation}
\label{sec:supp_audit_eqprop}

\noindent
\textbf{Reference.} Scellier \& Bengio, ``Equilibrium Propagation: Bridging the Gap between Energy-Based Models and Backpropagation,'' Frontiers in Computational Neuroscience 11:24, 2017.

\noindent
\textbf{Placement (dual reading).}
\begin{itemize}[leftmargin=*, itemsep=2pt]
  \item \textbf{Primary:} Continuous-time / Approximate or implicit gradient; Functional analogy; Local or algorithmic (mechanistic-plasticity reading on the learning axis is available).
  \item \textbf{Secondary:} Implicit / Energy or unsupervised. Reading the state-dynamics class as fixed-point of the energy function (rather than continuous trajectory) and the credit-assignment row as energy-based learning is internally consistent but de-emphasizes the algorithm's defining contribution.
\end{itemize}

\noindent
\textbf{Rationale.} The state evolves under a continuous-time gradient dynamics relaxing to an energy minimum; the two-phase free/nudged procedure estimates the loss gradient, with the gradient estimate exact in a particular limit and otherwise equivalent to recurrent backpropagation. The defining contribution is gradient estimation in a dynamical system, which motivates the primary placement in the approximate-gradient row. The second-phase weight change has been argued to correspond to a form of spike-timing-dependent plasticity, giving the learning grounding its mechanistic-plasticity reading. Symmetric weights are required.

\subsection{Forward-Forward (applied to MLP)}
\label{sec:supp_audit_ff}

\noindent
\textbf{Reference.} Hinton, ``The Forward-Forward Algorithm: Some Preliminary Investigations,'' NeurIPS 2022 presentation (arXiv:2212.13345).

\noindent
\textbf{Placement.} Static (when applied to an MLP; the state-dynamics class is application-dependent and inherits from the host architecture); Energy or unsupervised; None or historical inspiration; Local or algorithmic.

\noindent
\textbf{Rationale.} Each layer optimizes a local goodness function (sum of squared activations), raised for positive data and lowered for negative data, with no global error propagation. The stated motivation is the biological implausibility of backpropagation: there is no convincing evidence the cortex propagates error derivatives or stores activations for a backward pass, and the algorithm replaces the forward+backward pair with two forward passes using local goodness objectives. The motivation is therefore about the locality of the learning rule, not an architectural constraint, so forward grounding remains at None or historical while learning grounding sits at local or algorithmic. The row commitment to Energy or unsupervised is invariant across host architectures.

\subsection{Spiking neural network configurations}
\label{sec:supp_audit_snn}

The leaky integrate-and-fire forward architecture, with continuous subthreshold membrane dynamics and discrete spike events, directly adopts biological substrates from Section~2.1 of the main manuscript and sits at mechanistic-plausibility forward grounding regardless of the learning method. On the learning axis, the same forward architecture admits two configurations that differ entirely from each other, providing the clearest illustration in the audit of the forward and learning grounding split.

\subsubsection{Surrogate-gradient SNN}
\label{sec:supp_audit_snn_surrogate}

\noindent
\textbf{Reference.} Neftci, Mostafa \& Zenke, ``Surrogate Gradient Learning in Spiking Neural Networks,'' IEEE Signal Processing Magazine 36(6):51--63, 2019.

\noindent
\textbf{Placement.} Hybrid event-driven; Approximate or implicit gradient; Mechanistic plausibility; Weak.

\noindent
\textbf{Rationale.} The non-differentiable spike threshold is replaced with a smooth surrogate derivative during the backward pass, enabling backpropagation through the unrolled spike train. The surrogate is selected for optimization convenience (typical choices include fast-sigmoid or boxcar derivatives around threshold), not because any biological mechanism dictates its shape; no specific synaptic-plasticity mechanism is invoked to justify the surrogate. The learning grounding is therefore Weak despite the strongly grounded forward dynamics.

\subsubsection{STDP-trained SNN}
\label{sec:supp_audit_snn_stdp}

\noindent
\textbf{References.} Bi \& Poo, ``Synaptic modifications in cultured hippocampal neurons: dependence on spike timing, synaptic strength, and postsynaptic cell type,'' J.\ Neuroscience 18(24):10464--10472, 1998; Markram, L\"ubke, Frotscher \& Sakmann, ``Regulation of synaptic efficacy by coincidence of postsynaptic APs and EPSPs,'' Science 275:213--215, 1997.

\noindent
\textbf{Placement.} Hybrid event-driven; Local plasticity; Mechanistic plausibility; Mechanistic plasticity.

\noindent
\textbf{Rationale.} The learning rule is the empirically measured plasticity mechanism: pre-before-post spiking within a tens-of-millisecond window yields long-term potentiation, and the reverse yields depression, with weight change approximately exponential in the spike-time difference. Both forward and learning axes are anchored in biophysical measurement, satisfying the strict mechanistic-plasticity criterion. \emph{Limited evidence annotation.} Deep STDP-only networks at modern scale remain thin: STDP-trained SNNs have not been demonstrated competitively beyond small and medium image benchmarks, and most deep SNN results at ImageNet scale use surrogate gradients or ANN-to-SNN conversion rather than pure STDP.

\subsection{Transformer}
\label{sec:supp_audit_transformer}

\noindent
\textbf{Reference.} Vaswani et al., ``Attention Is All You Need,'' NeurIPS 2017.

\noindent
\textbf{Placement.} Discrete-time / sequence (sequence-position indexing via positional encodings; no persistent hidden-state recurrence during parallel training); Global gradient; None or historical inspiration; None.

\noindent
\textbf{Rationale.} Recurrence and convolution are dispensed with; sequence order is modeled through positional encodings and parallel self-attention. Neither the architecture nor the training method invokes a biological constraint. The autoregressive KV cache used at inference is an inference-time optimization rather than a separate configuration.

\subsection{Ambiguous-Classification Audit Table}
\label{sec:supp_audit_table}

Configurations with a dual reading (modern Hopfield, BP-equivalent predictive coding, equilibrium propagation) appear in the consolidated table below under their primary placement; the secondary reading is in the corresponding body subsection of Section~\ref{sec:supp_audit}.

\begin{table*}[h!]
\centering
\small
\begin{tabularx}{\textwidth}{@{}p{4.2cm} p{3.0cm} p{2.6cm} p{2.3cm} X@{}}
\toprule
\textbf{Configuration} & \textbf{State-dynamics class} & \textbf{Learning row} & \textbf{Forward grounding} & \textbf{Learning grounding} \\
\midrule
Modern Hopfield (end-to-end) & Implicit / fixed-point & Global gradient & Functional analogy & None \\
Classical Hopfield (Hebbian) & Discrete (fixed-point fn.) & Local plasticity & Arch.\ constraint & Local or algorithmic \\
PC: recurrent inference & Continuous-time / fixed-pt. & Local plasticity & Arch.\ constraint & Local/algorithmic \\
PC: energy-based & Continuous-time / fixed-pt. & Energy/unsupervised & Arch.\ constraint & Local/algorithmic \\
PC: local-error & Implicit / fixed-point & Approx./implicit & Funct.\ analogy & Local/algorithmic \\
PC: BP-equivalent limit & Implicit / fixed-point & Approx./implicit & Funct.\ analogy & Local/algorithmic \\
Equilibrium propagation & Continuous-time & Approx./implicit & Funct.\ analogy & Local/algorithmic \\
Forward-Forward (on MLP) & Static & Energy/unsupervised & None/Hist. & Local/algorithmic \\
Surrogate-gradient SNN & Hybrid event-driven & Approx./implicit & Mech.\ plausibility & Weak \\
STDP-trained SNN & Hybrid event-driven & Local plasticity & Mech.\ plausibility & Mech.\ plasticity \\
Transformer & Discrete / sequence & Global gradient & None/Hist. & None \\
\bottomrule
\end{tabularx}

\vspace{0.3em}
\noindent\footnotesize Detailed placement rationales for each row are provided in the corresponding body subsection of Section~\ref{sec:supp_audit}.
\end{table*}

\section{Configuration Ledger for Figure~2}
\label{sec:supp_ledger}

The following ledger enumerates every architecture--learning configuration counted in Figure~2 of the main manuscript. Each row records a unique configuration ID, the architecture, the learning rule, the state-dynamics class (per Section~3.1 of the main manuscript), the credit-assignment row (Section~3.2), the forward and learning biological-grounding tiers (Section~3.3), and a compact evidence pointer. Aggregating the ledger yields exactly the ribbon widths and node heights of Figure~2: 32 configurations distributed as Static~6, Discrete-time~14, Continuous-time~6, Implicit~4, Hybrid~2 on the forward axis, and Global-gradient~16, Approximate-or-implicit~7, Local~6, Energy/unsupervised~3 on the credit-assignment axis. Configurations flagged in Section~\ref{sec:supp_audit} contribute to the ledger under their primary placement.

\begin{table*}[h!]
\centering
\scriptsize
\setlength{\tabcolsep}{3pt}
\begin{tabularx}{\textwidth}{@{}p{0.5cm} p{2.6cm} p{2.6cm} p{1.8cm} p{1.9cm} p{1.5cm} p{1.5cm} X@{}}
\toprule
\textbf{ID} & \textbf{Architecture} & \textbf{Learning rule} & \textbf{Forward class} & \textbf{Credit class} & \textbf{Fwd.\ gnd.} & \textbf{Learn.\ gnd.} & \textbf{Evidence} \\
\midrule
C01 & MLP & Backpropagation & Static & Global grad. & None/Hist. & None & \cite{rumelhart1986learning} \\
C02 & CNN (LeNet, AlexNet) & Backpropagation & Static & Global grad. & None/Hist. & None & \cite{lecun1998gradient,krizhevsky2012imagenet} \\
C03 & ResNet & Backpropagation & Static & Global grad. & None/Hist. & None & \cite{he2016deep} \\
C04 & PINN & Backpropagation (with PDE-residual objective) & Static & Global grad. & None/Hist. & None & \cite{raissi2019physics} \\
C05 & Neocognitron & Local competitive Hebbian & Static & Local plast. & Arch.\ constr. & Local/alg. & \cite{fukushima1980neocognitron} \\
C06 & MLP + Forward-Forward & Layerwise goodness & Static & Energy/unsup. & None/Hist. & Local/alg. & \cite{hinton2022forward} \\
\midrule
C07 & Simple RNN & BPTT & Discrete & Global grad. & Funct.\ analogy & None & \cite{elman1990finding,werbos1990backpropagation} \\
C08 & LSTM & BPTT & Discrete & Global grad. & Funct.\ analogy & None & \cite{hochreiter1997long} \\
C09 & GRU & BPTT & Discrete & Global grad. & Funct.\ analogy & None & \cite{cho2014learning} \\
C10 & Transformer & Backpropagation & Discrete & Global grad. & None/Hist. & None & \cite{vaswani2017attention} \\
C11 & S4 & Backpropagation & Discrete & Global grad. & None/Hist. & None & \cite{gu2022efficiently} \\
C12 & Mamba & Backpropagation & Discrete & Global grad. & None/Hist. & None & \cite{gu2023mamba} \\
C13 & RWKV & Backpropagation & Discrete & Global grad. & None/Hist. & None & \cite{peng2023rwkv} \\
C14 & xLSTM & Backpropagation & Discrete & Global grad. & None/Hist. & None & \cite{beck2024xlstm} \\
C15 & RNN + synthetic gradients & Decoupled interface & Discrete & Approx./impl. & Funct.\ analogy & Weak & \cite{jaderberg2017decoupled} \\
C16 & RNN + random-feedback local online (RFLO) & Local eligibility + random feedback & Discrete & Approx./impl. & Funct.\ analogy & Weak & \cite{murray2019local} \\
C17 & Classical Hopfield & Hebbian storage & Discrete & Local plast. & Arch.\ constr. & Local/alg. & \cite{hopfield1982neural,amit1985storing} \\
C18 & RBM & Contrastive divergence & Discrete & Energy/unsup. & None/Hist. & Local/alg. & \cite{hinton2002training} \\
C19 & DBN & Contrastive divergence & Discrete & Energy/unsup. & None/Hist. & Local/alg. & \cite{hinton2006fast} \\
C32 & HTM & Dendrite-local Hebbian & Discrete & Local plast. & Arch.\ constr. & Local/alg. & \cite{hawkins2016neurons,hawkins2019framework} \\
\midrule
C20 & CTRNN & Solver BP / BPTT & Continuous & Global grad. & Funct.\ analogy & None & \cite{pearlmutter1989learning} \\
C21 & Liquid Time-Constant NN & Solver BP & Continuous & Global grad. & Funct.\ analogy & None & \cite{hasani2021liquid} \\
C22 & Liquid closed-form & BPTT / reverse-mode through closed-form recurrence & Continuous & Global grad. & Funct.\ analogy & None & \cite{hasani2022closed} \\
C23 & Neural ODE & Adjoint method & Continuous & Approx./impl. & None/Hist. & None & \cite{chen2018neural} \\
C24 & PC: recurrent inference & Local error min. & Continuous & Local plast. & Arch.\ constr. & Local/alg. & \cite{rao1999predictive} \\
C25 & CT-RNN + reward-mod.\ Hebbian & Reward-modulated Hebbian & Continuous & Local plast. & Funct.\ analogy & Local/alg. & \cite{miconi2017biologically} \\
\midrule
C26 & Modern Hopfield & End-to-end BP & Implicit & Global grad. & Funct.\ analogy & None & \cite{ramsauer2021hopfield} \\
C27 & DEQ & Implicit differentiation & Implicit & Approx./impl. & None/Hist. & None & \cite{bai2019deep} \\
C28 & PC: local-error & Approx.\ BP via local errors & Implicit & Approx./impl. & Funct.\ analogy & Local/alg. & \cite{whittington2017approximation} \\
C29 & PC: BP-equivalent limit & Local errors, BP-equivalent & Implicit & Approx./impl. & Funct.\ analogy & Local/alg. & \cite{millidge2022predictive,song2020can} \\
\midrule
C30 & LIF SNN & Surrogate gradient & Hybrid & Approx./impl. & Mech.\ plaus. & Weak & \cite{neftci2019surrogate} \\
C31 & LIF SNN & STDP & Hybrid & Local plast. & Mech.\ plaus. & Mech.\ plast. & \cite{bi1998synaptic,markram1997regulation} \\
\bottomrule
\end{tabularx}
\end{table*}

\noindent
\textbf{Reconciliation.} Aggregating the ledger by forward class yields Static 6 (C01--C06), Discrete-time 14 (C07--C19, C32), Continuous-time 6 (C20--C25), Implicit 4 (C26--C29), Hybrid 2 (C30--C31), matching the left-column node heights of Figure~2. Aggregating by credit-assignment class yields Global gradient 16 (C01--C04, C07--C14, C20--C22, C26), Approximate/implicit 7 (C15, C16, C23, C27, C28, C29, C30), Local plasticity 6 (C05, C17, C24, C25, C31, C32), Energy/unsupervised 3 (C06, C18, C19), matching the right-column node heights. Ribbon widths in Figure~2 correspond to cell counts of the joint (forward, credit) distribution and are recoverable from the ledger by grouping. Section~\ref{sec:supp_audit} adjudicates a broader set of ambiguous cases than are counted here: some entries in that audit (modern Hopfield: C26; predictive coding in its BP-equivalent limit: C29) are counted rows of this ledger, while others (equilibrium propagation, additional predictive-coding variants) are adjudicated exemplars used to support classification decisions but are not themselves added as new counted configurations. Figure~2 uses the declared primary placements throughout; the alternative readings documented in Section~\ref{sec:supp_audit} concern biological-grounding tier assignment or interpretive framing rather than state-dynamics/credit-assignment coordinates and therefore do not move ribbons in Figure~2.

\textbf{Reservoir exclusion.} Echo-state networks (ESN) and liquid state machines (LSM), which appear in the model-family table in Section~3.3 of the main manuscript, are omitted from the Figure~2 ledger because they train only a linear readout on top of a fixed random reservoir; the internal recurrent dynamics do not participate in credit assignment. A configuration counted in Figure~2 requires the learning rule to update parameters that shape the forward state dynamics, which excludes fixed-reservoir readout-only training on principled rather than incidental grounds.

\textbf{Mechanism-level sensitivity of the backward axis.} A reviewer concern is whether the Global-gradient dominance in Figure~2 is an artifact of bundling several distinct gradient procedures into one category. Table~\ref{tab:supp_mechanism_decomposition} decomposes the two largest backward categories by the specific credit-assignment mechanism each ledger row uses. Ordinary reverse-mode backpropagation alone accounts for 10 of 32 configurations, which is larger than the entire Approximate/implicit category (7 of 32). The Approximate/implicit category fragments across seven distinct mechanisms, each contributing exactly one configuration. Refining the backward axis by actual mechanism preserves the asymmetry: the visual dominance of Global gradient in Figure~2 is not a bundling artifact but reflects the concentration of audited configurations in a single canonical mechanism, while the alternatives are individually small even after being counted separately.

\begin{table}[!htbp]
\centering
\caption{Mechanism-level decomposition of the two largest backward categories. Each mechanism is identified by the exact procedure it uses, and the count reports how many of the 32 ledger configurations it accounts for. Global gradient (16 configurations total) is dominated by ordinary reverse-mode backpropagation (10); the Approximate/implicit category (7 configurations total) fragments across seven distinct mechanisms, each contributing exactly one configuration.}
\label{tab:supp_mechanism_decomposition}
\small
\setlength{\tabcolsep}{6pt}
\renewcommand{\arraystretch}{1.10}
\begin{tabular}{@{}l l r l@{}}
\toprule
\textbf{Backward category} & \textbf{Mechanism} & \textbf{Count} & \textbf{Ledger IDs} \\
\midrule
Global gradient (16)      & Ordinary reverse-mode BP        & 10 & C01--C04, C10--C14, C26 \\
                          & BPTT / reverse-mode through recurrence & 4 & C07--C09, C22 \\
                          & Solver BP / solver BPTT         & 2 & C20--C21 \\
\midrule
Approximate/implicit (7)  & Synthetic gradients (DNI)       & 1 & C15 \\
                          & Random-feedback local online (RFLO) & 1 & C16 \\
                          & Adjoint sensitivity             & 1 & C23 \\
                          & Implicit differentiation        & 1 & C27 \\
                          & Approx.\ BP via local errors (PC) & 1 & C28 \\
                          & Local errors, BP-equivalent limit (PC) & 1 & C29 \\
                          & Surrogate gradient              & 1 & C30 \\
\bottomrule
\end{tabular}
\end{table}

\section{Empirical Evidence for the Scaling Claim}
\label{sec:supp_scaling}
\label{sec:supp_evidence_scaling}

The main manuscript states an evidence-bounded version of the scaling claim: the largest scales remain concentrated in global or closely gradient-derived and error-propagation mechanisms, while strictly local and mechanistic-plasticity rules remain substantially less scalable. This appendix records the strongest reported per-family results that anchor the claim, including recent counterexamples that narrow but do not close the gap. The cutoff is the literature search date in Section~\ref{sec:supp_litsearch} above.

\begin{center}
\small
\begin{longtable}{@{}p{3.4cm} p{5.0cm} p{2.4cm} p{2.4cm}@{}}
\caption{Empirical scaling evidence anchoring the main paper's scaling claim: strongest reported per-family results at the search cutoff, and their comparison against mainstream backpropagation.}
\label{tab:supp_scaling_evidence} \\
\toprule
\textbf{Family} & \textbf{Largest reported result} & \textbf{Comparison} & \textbf{Reference} \\
\midrule
\endfirsthead
\multicolumn{4}{@{}l}{\textit{Table~\ref{tab:supp_scaling_evidence} continued from previous page}} \\
\toprule
\textbf{Family} & \textbf{Largest reported result} & \textbf{Comparison} & \textbf{Reference} \\
\midrule
\endhead
\midrule
\multicolumn{4}{r@{}}{\textit{continued on next page}} \\
\endfoot
\bottomrule
\endlastfoot
Feedback alignment (plain) & Performs well on MNIST/CIFAR; significantly below BP on ImageNet & Gap to BP large at ImageNet & \cite{bartunov2018assessing} \\
Sign-symmetry (FA variant) & Approaches BP on ImageNet with ResNet-18/AlexNet & Within a few percent of BP & \cite{xiao2018sign} \\
Target propagation & MNIST/CIFAR competitive; ImageNet substantially below BP & No parity at ImageNet scale & \cite{bartunov2018assessing} \\
Equilibrium propagation & 11.7\% test error on CIFAR-10 with deep ConvNets; approaches BPTT to within roughly 0.6 percentage points; downsampled ImageNet 32$\times$32 in subsequent work & No full ImageNet-1k or LM result & \cite{laborieux2021scaling,laborieux2022holomorphic} \\
Weight Mirror (weight-symmetry relaxation) & Learns dynamic feedback weights to align with forward weights; on ImageNet, ResNet-18 top-1 error 30.2\% for Weight Mirror vs 30.1\% for BP (within 0.1 pp), ResNet-50 top-1 error 23.4\% for Weight Mirror vs 22.9\% for BP (within 0.5 pp); the authors describe the result as ``nearly matched backprop'' and ``learning ImageNet about as well as backprop does'' & Approximately matches BP on ResNet-18 and ResNet-50 ImageNet under the paper's training pipeline; substantially outperforms plain feedback alignment and sign-symmetry; still relies on error-broadcast machinery & \cite{akrout2019deep} \\
Product Feedback Alignment (PFA) & On ResNet-18 ImageNet: BP 69.69\%, PFA 68.46\%, PFA-o 69.30\% top-1 accuracy; approaches BP without explicit weight symmetry & PFA-o within roughly 0.4 percentage points of BP at ResNet-18 scale, but no ResNet-50 or GPT-class demonstration reported & \cite{li2024deep} \\
Predictive coding & Matches BP on small ConvNets on CIFAR-10; recent regularization extends to roughly 15-layer Tiny-ImageNet models approaching BP; BP-equivalent limit matches on SVHN/CIFAR at $\approx 100\times$ compute & No full ImageNet-1k or LM result; large compute overhead & \cite{salvatori2025neuromimetic,qi2025towards,millidge2022predictive} \\
Forward-Forward (FF) & First successful FF training on ImageNet (ASGE variant): Top-1 26.21\%, Top-5 47.49\% & Standard ResNet-50 BP baseline: Top-1 76.21\%, Top-5 92.97\% (Bag-of-Tricks training) & \cite{asge2025}; baseline \cite{he2019bagoftricks} \\
Dendritic-local learning & DLL trains ANNs, CNNs, ResNet, and Transformers with a spatially and temporally local, dendrite-inspired plasticity rule; competitive with BP on the reported CIFAR-10, CIFAR-100, and small-ImageNet-subset benchmarks & No full ImageNet-1k or GPT-class demonstration; scale reported is below ResNet-50 ImageNet & \cite{lv2025dendritic} \\
Directly trained SNN (surrogate gradient, spiking Transformer) & QKFormer reaches 85.65\% top-1 on ImageNet-1k at 64.96M parameters (spiking self-attention with Q-K token attention) & Comparable to modest ANN Transformers; still trained with surrogate-gradient backpropagation (not a biologically constrained learning rule) & \cite{zhou2024qkformer} \\
\end{longtable}
\end{center}

\subsection*{Operational definitions used in the scaling claim}

The scaling claim is bounded by the following operational definitions:
\begin{itemize}[leftmargin=*, itemsep=2pt]
  \item ``Strictly local or mechanistic-plasticity rule'' means one that does not propagate an error signal through the forward graph in any form, whether transported (BP), randomized (FA), aligned (Weight Mirror), or product-decomposed (PFA); Hebbian learning, STDP, dendritic local rules, and eligibility-trace rules with locally computed learning signals fall in this category.
  \item ``Error-broadcast approximate-gradient rule'' means one that relaxes weight transport, update locking, or exact-gradient computation but still routes a global or layerwise error signal back to earlier layers. Feedback alignment, direct FA, sign-symmetric variants, Weight Mirror, and PFA fall in this category.
  \item ``Mainstream backpropagation'' means standard backpropagation (through time when applicable) on a contemporary backbone (ResNet, Transformer, or comparable), trained with conventional optimizers.
  \item ``The frontier'' is operationalized as full-resolution ImageNet-1k classification at ResNet-50 level, GPT-class autoregressive language modeling, or standard reinforcement-learning benchmarks at the scale reported in published baselines through the search cutoff date.
\end{itemize}

\subsection*{Falsifying threshold}

The scaling claim as stated would be revised by a demonstration in which a strictly local or mechanistic-plasticity learning rule (i.e., one not relying on backpropagation-style error signals through the forward graph, whether transported or randomized) reaches backpropagation-parity accuracy at full ImageNet-1k or at a GPT-class language model with comparable training compute. As of the search cutoff, no such demonstration has been published, though several works have narrowed the gap using approximate-gradient mechanisms that relax weight transport or update locking while retaining an error-signal broadcast.

\section{Audit of Apparently Sparse Regions of the Joint Taxonomy}
\label{sec:supp_sparse_audit}

Four apparently sparse (state-dynamics, credit-assignment) combinations in Figure~2 could indicate research gaps, conceptual incompatibilities, or taxonomy limitations. This appendix audits each and records the determination.

\subsection{Implicit / Local plasticity}
\label{sec:supp_sparse_implicit_local}

\textbf{Determination: occupied, not a structural gap.} The least-control principle \cite{meulemans2022least} provides a direct occupant by training equilibrium/implicit-state systems with a spatially and temporally local rule. Equilibrium propagation \cite{scellier2017equilibrium} is an adjacent ambiguous case; under the primary placement adopted in Section~\ref{sec:supp_audit_eqprop}, it remains Continuous-time / Approximate or implicit gradient.

\subsection{Continuous-time / Energy or unsupervised}
\label{sec:supp_sparse_ctode_energy}

\textbf{Determination: near-empty in the audited literature; candidate research gap.} The audit found no clearly published work combining a true non-converging neural-ODE trajectory (in the style of Chen et al.\ 2018) with an energy-based or Forward-Forward-style unsupervised objective. Standard Neural ODEs use the adjoint method (approximate gradient row); energy-based methods typically operate on discrete-time recurrence or implicit fixed points. The continuous-time energy combination is flagged as a candidate research direction.

\subsection{Continuous-time / Local plasticity (excluding spiking with STDP)}
\label{sec:supp_sparse_ctode_local}

\textbf{Determination: occupied, not a gap, but limited evidence at scale.} Miconi \cite{miconi2017biologically} trains continuous-time rate recurrent networks with a reward-modulated Hebbian local rule (non-spiking, non-STDP). Hoerzer, Legenstein and Maass \cite{hoerzer2014emergence} demonstrate a related reward-modulated Hebbian rule in rate recurrent networks. Demonstrated scale is small (cognitive tasks at hundreds of units); the limitation is on scale rather than on existence.

\subsection{Static / Approximate or implicit gradient}
\label{sec:supp_sparse_static_approx}

\textbf{Determination: occupied, not a gap.} PEPITA \cite{dellaferrera2022pepita} and Direct Feedback Alignment \cite{nokland2016direct} are static feedforward networks trained with approximate or implicit gradients (a second forward pass with random feedback modulation in PEPITA; fixed random direct feedback in DFA). Both have demonstrated competitive performance on small to medium classification benchmarks.

\subsection*{Net finding}

Three of the four flagged combinations are occupied once works are reclassified along the correct axis of the joint taxonomy. Only the continuous-time non-equilibrium plus energy/unsupervised combination remained near-empty in the audited literature.

\bibliographystyle{ACM-Reference-Format}
\bibliography{references}